\documentclass[aps,prb,twocolumn,amsmath,amssymb,showpacs]{revtex4-1}
\usepackage{graphicx}
\usepackage{graphics}
\usepackage{amsfonts,epsfig}
\usepackage{float}
\usepackage{color}
\allowdisplaybreaks
\usepackage{multirow, makecell}
\usepackage{subcaption}
\usepackage{booktabs}
\usepackage{bm}
\usepackage{tabulary}
\usepackage{hhline}
\usepackage{multirow, rotating}
\usepackage{threeparttable}
\usepackage{amsfonts}
\usepackage{caption} 
\usepackage{graphicx}
\usepackage{epsfig}

\newcommand{\Transpose}{^{\mathsf{T}}}

\newcommand{\HiddenStateSymb}{h}
\newcommand{\HiddenState}{\bm{\HiddenStateSymb}}

\newcommand{\HiddenStatexSymb}{k}
\newcommand{\HiddenStatex}{\bm{\HiddenStatexSymb}}

\newcommand{\HiddenDim}[1]{n_{#1}}

\newcommand{\LeftSingularVectorSymb}{S}
\newcommand{\LeftSingularVector}{\LeftSingularVectorSymb}

\newcommand{\RightSingularVectorSymb}{D}
\newcommand{\RightSingularVector}{\RightSingularVectorSymb}

\newcommand{\SingularValueSymb}{V}
\newcommand{\SingularValue}{\SingularValueSymb}

\begin{document}

\title{Constants of motion network revisited}

\author{Wenqi Fang}
\email{wq.fang@siat.ac.cn}
\author{Chao Chen, Yongkui Yang and Zheng Wang}
 \affiliation{Shenzhen Institute of Advanced Technology, Chinese Academy of Sciences, 
Shenzhen, 518055, Guangdong, China}

\date{\today}

\begin{abstract}
  Discovering constants of motion is meaningful in helping understand the dynamical systems, but inevitably needs proficient mathematical skills and keen analytical capabilities. With the prevalence of deep learning, methods employing neural networks, such as Constant Of Motion nETwork (COMET), are promising in handling this scientific problem. Although the COMET method can produce better predictions on dynamics by exploiting the discovered constants of motion, there is still plenty of room to sharpen it. In this paper, we propose a novel neural network architecture, built using the singular-value-decomposition (SVD) technique, and a two-phase training algorithm to improve the performance of COMET. Extensive experiments show that our approach not only retains the advantages of COMET, such as applying to non-Hamiltonian systems and indicating the number of constants of motion, but also can be more lightweight and noise-robust than COMET. 
\end{abstract}


\maketitle

\section{Introduction}

The law of conservation is one of the basic physical laws of the world.
It is essentially governed by Noether's theorem~\cite{noether1971invariant}, i.e. there is a one-to-one correspondence between conserved quantities of the system and its symmetry. As it turns out, conserved quantities, or called constants of motion, generally play a more important role than superficial dynamics in understanding the physical world \cite{anderson1972more, elesedy2023symmetry, sillerud2024space}. However, symmetries and constants of motion of the dynamical system to be solved are usually too subtle to be discovered, and thus cannot be exploited effectively. Therefore, developing efficient and practical approaches to discover hidden symmetries or constants of motion is extremely necessary to help in-depth understanding and problem-solving. 

Historically, researchers tried to discover the hidden constants of motion through time-consuming hand-crafted strategies, performing multifarious analytical and numerical works based on observational data or mathematical descriptions for each specific problem.
For example, Daniel Bernoulli and Leonhard Euler independently proposed the law of conservation of angular momentum after thoroughly analyzing Kepler's Second Law of planetary motion discovered about one hundred years ago~\cite{sparavigna2015historical}. 
With the recent rapid development of deep learning, employing neural networks for scientific machine learning from observational data is trending \cite{mukhamediev2021classical, subramanian2024towards}. 
Among them, there is a large body of literature about discovering constants of motion in the dynamical systems, such as Hamiltonian-based neural network and its variants \cite{greydanus2019hamiltonian, chen2021nsf, cranmer2020lnn, jin2022learning-poisson, zhang2024learning}, COMET \cite{kasim2022constants}, FINDE \cite{matsubara2022finde}, ConservNet \cite{ha2021discovering} and ConCerNet \cite{zhang2023concernet} etc. Inspired by these methods, after a critical analysis of the COMET method, this paper attempts to sharpen it from the following two aspects.

Firstly, COMET utilizes common multilayer perceptrons to help represent unknown variables. However, since the chosen architecture of neural networks is apt to be overparameterized in the machine learning field \cite{cooper2018loss, adeoye2024regularized}, it is irresistible to wonder if there is a more lightweight way to construct neural networks when solving the constants of motion. Following this line of thinking, we propose to adopt singular-value-decomposition (SVD)-like matrix decomposition to deal with the weight parameters of COMET, thus reducing the number of parameters by a large margin. This technique is also known elsewhere as the low-rank method \cite{cho2024hypernetwork} or reduced-order modeling \cite{barraza2024reduced}. 
Moreover, to accommodate this SVD-like structure of neural networks, we develop a two-phase training algorithm to train them.

Secondly, in the field of using deep learning methods to find the constants of motion,  there is less discussion about the noise robustness of the used methods. To fill this gap, we focus on analyzing whether our method is more robust to the noise compared to the COMET method. Due to the introduced SVD trick, we spontaneously consider our method to be noise-robust \cite{epps2019singular}. As it turns out, extensive experiments on various typical dynamical systems demonstrate our proposed method indeed has this superiority over the COMET method.

According to the discussion above, based on the COMET method, we propose a novel approach, named Meta-COMET, to discover the hidden constants of motion in dynamical systems. It turns out that our method is more lightweight and noise-robust than COMET. The main contributions of this paper are threefold:

1. Inspired by the low-rank method in physics-informed neural network (PINN) \cite{cho2024hypernetwork}, we meticulously design a neural network architecture employing the SVD method to significantly reduce the number of parameters in the neural networks of COMET, for example by more than 14 times in the mass-spring case, without impairing its performance.

2. Different from the traditional training method in COMET, a two-phase training algorithm is introduced in Meta-COMET to adapt to the SVD-like neural network architecture. This two-phase training algorithm ensures that specific weight parameters in the neural network architecture satisfy the semi-orthogonal constraints in SVD. 

3. Meta-COMET is validated on a wide range of dynamical systems, ranging from toy systems, such as mass-spring, to complex systems that are learned from pixel training data. Despite using lightweight neural network architecture, our method outperforms COMET in terms of noise robustness.

\section{Related works}

Learning the dynamics of systems is an ancient but timeless problem since it is closely related to much of science and engineering \cite{forrester2007system}. 
With the resurgence of deep learning, more and more researchers are using neural networks to dig into this classical topic instead of traditional manual numerical or analytical ways \cite{tiumentsev2019neural}.

Neural ordinary differential equation (NODE) is a famous and prevalent method to analyze dynamical systems by solving the ODE equations \cite{chen2018neural-ode}.
However, NODE lacks appropriate encoding of inductive biases about the conserved quantities in systems. Therefore, it often struggles to discover the constants of motion. 

Unlike NODE, Hamiltonian neural network (HNN) utilized Hamiltonian mechanics to learn exact conservation laws, i.e. energy conservation, in dynamical systems in an unsupervised manner \cite{greydanus2019hamiltonian}. Due to its elegance and simplicity, HNN was subsequently extended or modified to dissipative HNN \cite{greydanus2022dissipative-hnn, zhong2020dissipative}, adaptable HNN \cite{han2021adaptable},  constrained HNN \cite{finzi2020simplifying}, Hamiltonian neural Koopman operator\cite{zhang2024learning}, HNN with symplectic integrator \cite{chen2019symplectic-rnn, zhong2019sympnet} etc. Unfortunately, because of its intrinsic limitation, a Hamiltonian-based neural network only works using canonical coordinates instead of arbitrary coordinates. 

Several works tried to relax this canonical coordinates constraint in HNN and developed a few coordinate-free methods, such as 
Neural symplectic form \cite{chen2021neural}, 
Lagrangian neural networks \cite{cranmer2020lnn, lutter2019delan}, Poisson neural network \cite{jin2022learning-poisson}, etc. However, those works are still not flexible enough to adapt to non-Hamiltonian or non-Poisson systems.

Recently, other promising methods sprung up. For example, COMET utilized QR decomposition to ensure the state dynamics perpendicular to the gradient of each constant of motion \cite{kasim2022constants} and train it in a similar way to the PINN method \cite{ryu2024physics}. In the FINDE method, authors combined Lagrange multipliers and projection method to find the first integrals, meanwhile adopting HNN or NODE as the base model \cite{matsubara2022finde}. In addition, a few contrastive learning-based frameworks, such as ConservNet \cite{ha2021discovering} or ConCerNet \cite{zhang2023concernet}, were put forward to capture the invariants of the dynamical system automatically. Meanwhile, these two methods are among the few works that discuss the impact of noise on the discovery of constants of motion. However, the ConservNet method only modeled constants of motion, rather than states and constants of motion simultaneously \cite{ha2021discovering}.

\section{Preliminaries}
The COMET method deeply inspires our approach to discovering the constants of motion. Therefore, in this section, we review the key ideas of the COMET method. Other detailed information can be found in its original paper. 

For a common dynamical system, it can be solely represented by a set of its intrinsic states  $\mathbf{s}\in\mathbb{R}^{n_s}$, where $n_s$ is the number of states. The change of states describes the evolution of the system, and it typically depends on the states itself,  i.e. $d\mathbf{s}/dt = \mathbf{\dot{s}}(\mathbf{s})$, if there is no 
external force $\mathbf{F}$.

Unlike the dynamical property of states, the constants of motion, $\mathbf{c}\in\mathbb{R}^{n_c}$ with $n_c$ is the number of constant of motion, remain the same as their initial status. They are generally closely related to the states $\mathbf{s}$, and the property of the constants of motion can be described as follows:
\begin{equation}
\label{eq:constant-of-motion}
    \frac{d\mathbf{c}}{dt} = \frac{\partial \mathbf{c}}{\partial \mathbf{s}}\mathbf{\dot{s}} = \mathbf{0},
\end{equation}
where  $\partial \mathbf{c} / \partial\mathbf{s}$ is a $n_c\times n_s$ Jacobian matrix. Each row of the matrix is the gradient of each constant of motion with respect to the states $\mathbf{s}$. It implies that the gradient of each constant of motion and the state dynamics $\mathbf{\dot{s}}$ must be perpendicular to each other. 

According to this keen insight, the COMET method utilizes an orthogonalization process to solve the constants of motion in a dynamical system. To ensure the equation~(\ref{eq:constant-of-motion}) hold, it introduces QR decomposition, i.e.
\begin{align}
&    \mathbf{A} = \left(\nabla c_1, \nabla c_2, ..., \nabla c_{n_c}, \mathbf{\dot{s}_0}\right)  \nonumber \\
&   \mathbf{Q}, \mathbf{R} = \mathrm{QR}(\mathbf{A})  \\
    \label{eq:ortho-qr-method}
&    \mathbf{\dot{s}} = \mathbf{Q}_{(\cdot, n_c)} \mathbf{R}_{(n_c, n_c)},  \nonumber
\end{align}
where $c_i$ is the $i$-th element of the constants of motion $\mathbf{c}$, $\mathbf{\dot{s}_0}$ is the initial rate of change of the states. After concatenating them together, the obtained matrix $\mathbf{A}$ is a $n_s \times (n_c + 1)$ matrix. In addition, $\mathbf{Q}_{(\cdot, n_c)}$ is the last column of the matrix $\mathbf{Q}$, and $\mathbf{R}_{(n_c, n_c)}$ is the element at the last row and last column of the matrix $\mathbf{R}$. 
In this way, QR decomposition, implemented using Householder transformation \cite{householder1958unitary}, could find $n_c$ independent constants of motion and additionally impose an inequality constraint, i.e. $n_c < n_s$. Due to the unknown and arbitrary nature of $\mathbf{c}$ and $\mathbf{\dot{s}_0}$, COMET adopts deep neural networks to represent them.

To solve trainable functions $\mathbf{\dot{s}_0}(\mathbf{s})$ and $\mathbf{c}(\mathbf{s})$ simultaneously, the COMET method constructs the following loss function:
\begin{equation}
\label{eq:loss-function}
    \mathcal{L} = \left\lVert\mathbf{\dot{s}} -  \mathbf{\hat{\dot{s}}}\right\rVert^2 + w_1 \left\lVert\mathbf{\dot{s}_0} - \mathbf{\hat{\dot{s}}}\right\rVert^2 + w_2 \sum_{i=1}^{n_c} \left\lVert\nabla c_i \cdot \mathbf{\dot{s}_0}\right\rVert^2,
\end{equation}
where $\mathbf{\hat{\dot{s}}}$ is the training data  and $w_\cdot$ are the tunable regularization parameters. 
Furthermore, to make constraints~(\ref{eq:constant-of-motion}) to be fulfilled for the states outside the ones listed in the training dataset, extra noise is added to the last term in loss function $\mathcal{L}$. 
Due to this operation, the loss function $\mathcal{L}$ is represented as follows, i.e.
\begin{equation}
\begin{split}
    \mathcal{L} = & \left\lVert\mathbf{\dot{s}}(\mathbf{s}) -  \mathbf{\hat{\dot{s}}}\right\rVert^2 + w_1 \left\lVert\mathbf{\dot{s}_0}(\mathbf{s}) - \mathbf{\hat{\dot{s}}}\right\rVert^2 + \\
    & w_2 \sum_{i=1}^{n_c} \left\lVert\nabla c_i(\mathbf{s} + \mathbf{\Tilde{s}}) \cdot \mathbf{\dot{s}_0}(\mathbf{s} + \mathbf{\Tilde{s}})\right\rVert^2, 
\end{split}
\label{eq:loss-function-complete}
\end{equation}
where $\mathbf{\Tilde{s}}$ is the random noise sampled from a uniform distribution on the interval $[0, 1)$ with amplitude 0.1.
This loss function $\mathcal{L}$ is much like the 
PINN method where the first term of equation~(\ref{eq:loss-function-complete}) is the standard $L_2$ residual error, the second term of $\mathcal{L}$ is similar to the initial condition that needs to be satisfied, and the last term is the constraint equation imposed on the system to be solved \cite{ryu2024physics}.

\section{Methodology}
In this section, we detail our methodology, including a novel neural network architecture and a two-phase training algorithm, that can be used to improve the noise robustness capabilities of COMET in discovering the constants of motion in dynamic systems.


\subsection{Our neural network architecture}\label{sec:framework}
\begin{figure*}[t]
    \centering
    \includegraphics[width=\linewidth]{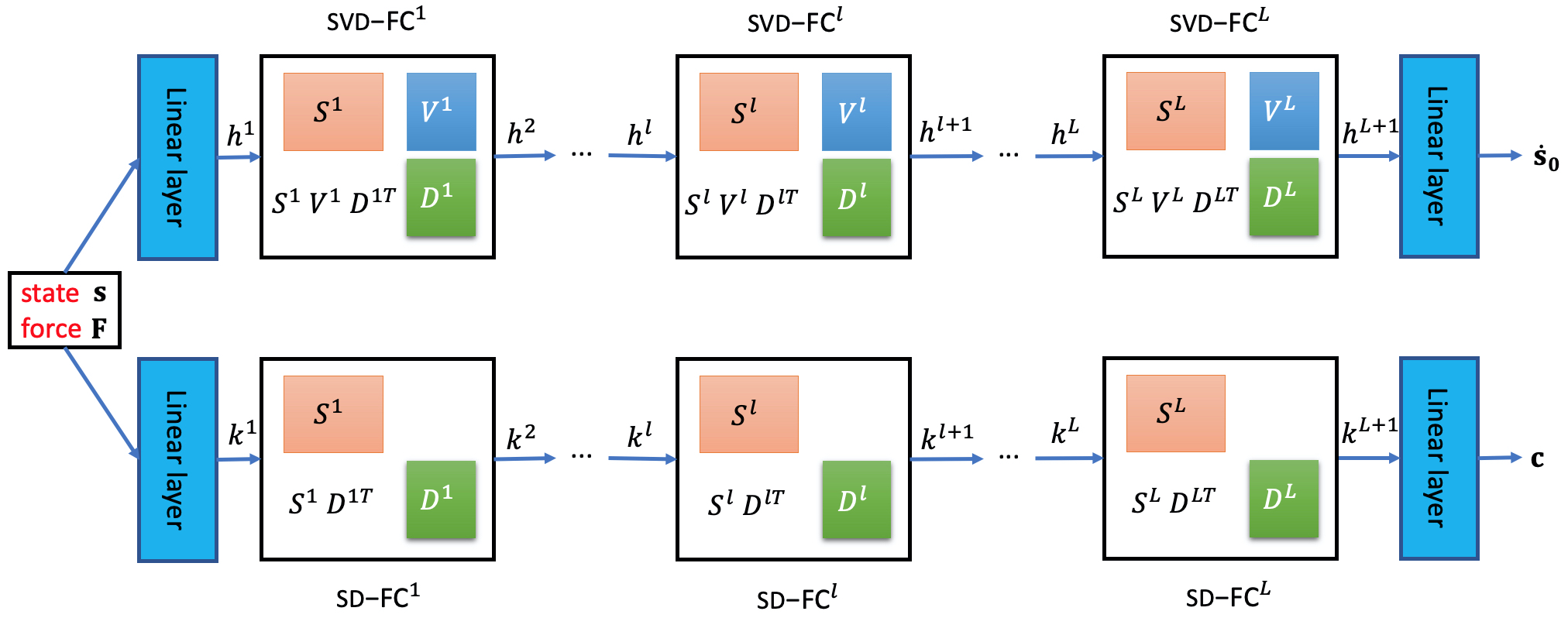}
    \caption{The architecture of the two neural networks to calculate $\mathbf{\dot{s}_0}$ and $\mathbf{c}$. The inputs of each neural network are both the concatenation of the state $\mathbf{s}$ and the external force $\mathbf{F}$ (if present).
    The parameters of the four linear layers are all different.
    In each layer $l$, except for the additional parameters $V^l$, the $\textsc{svd-FC}^l$ layer and the $\textsc{sd-FC}^l$ layer have exactly the same $S ^l$ and $D^l$.}
    \label{fig:framework}
\end{figure*}

In the COMET method,  the initial rate of change of the states $\mathbf{\dot{s}_0}$ and the constants of motion $\mathbf{c}$ are represented by one traditional fully connected (FC) neural network. 
This scheme ensures the inextricable correlation between these two variables and can calculate both of them at once. 
However, it inevitably introduces redundant parameters and is susceptible to noise to some extent \cite{cooper2018loss, adeoye2024regularized}.

In this paper, we resort to SVD trick to construct our neural network, shown in Figure~\ref{fig:framework}. 
There are two computational paths in our proposed architecture: 
one path for $\mathbf{\dot{s}_0}$ and the other path for $\mathbf{c}$. 
As shown in Figure~\ref{fig:framework}, the first and last layers are the common FC layers in both paths and the parameters of these layers are all different. However, the hidden layers are completely different from FC structure and are elaborately designed through the SVD technique.
We denote these intermediate layers in each path as \textsc{svd-FC} and \textsc{sd-FC} respectively. The relation between two consecutive hidden layers, $l$ and $l+1$, is defined as follows for each path: 
\begin{align}\label{eq:factored_fc}
    &  \HiddenState^{l+1} = \textsc{svd-FC}^{l}(\HiddenState^{l}) \  \Leftrightarrow \ \HiddenState^{l+1} = \LeftSingularVector^l(\SingularValue^l (\RightSingularVector^l{}\Transpose \HiddenState^l))   \\
    & \HiddenStatex^{l+1} = \textsc{sd-FC}^{l}(\HiddenStatex^{l}) \  \Leftrightarrow  \ \HiddenStatex^{l+1} = \LeftSingularVector^l(\RightSingularVector^l{}\Transpose \HiddenStatex^l),
\end{align}
where $\LeftSingularVector^l \in \mathbb{R}^{\HiddenDim{l+1} \times r}$ and $\RightSingularVector^l \in \mathbb{R}^{\HiddenDim{l} \times r}$ are full column-rank matrices (i.e., rank $r \ll n_l, n_{l+1}$), and $\SingularValue^l \in \mathbb{R}^{r \times r}$ is a diagonal matrix. 
In our framework, we set $\LeftSingularVector^l$, $\RightSingularVector^l$ matrixes in $\textsc{svd-FC}^{l}$ and $\textsc{sd-FC}^{l}$ are exactly the same in the corresponding $l$th layer. 
These settings indicate the variables, $\mathbf{\dot{s}_0}$ and $\mathbf{c}$, are inextricably linked to each other.

Both paths have the same input: the concatenation of state $\mathbf{s}$ and the force $\mathbf{F}$, depending on whether the system has an externally driven force. The internals to calculate $\mathbf{\dot{s}_0}$ can be described as with $\pmb{h}^{0} = \pmb{k}^{0} = [\mathbf{s}, \mathbf{F}]\Transpose$:
\begin{equation}\label{aaa}
    \begin{split}
&        \pmb{h}^1 = f(W_h^0 \pmb{h}^0 + \pmb{b_h}^0),\\
&        \pmb{h}^{l+1} = f(S^l ( ReLU(V^l) (D^l{}\Transpose \pmb{h}^l))), l=1,\ldots,L,\\
&       \mathbf{\dot{s}_0}(\mathbf{s}, \mathbf{F}) = W_h^{L+1} \pmb{h}^{L+1} + \pmb{b}_h^{L+1}, 
    \end{split}
\end{equation} 
where $f$ is the non-linear activation function between each layer, 
$\pmb{b}_.$ in these equations represent the bias parameters, and
the ReLU function is employed to automatically truncate the negative values so that the adaptive rank structure for the matrix $V^l$ can be achieved. This scheme can avoid setting different rank $r$ for each hidden layer to a certain extent.
Similarly, for the constants of motion $\mathbf{c}$, it can be expressed as follows: 
\begin{equation}\label{bbb}
    \begin{split}
&        \pmb{k}^1 = f(W_k^0 \pmb{k}^0 + \pmb{b}_k^0),\\
&        \pmb{k}^{l+1} = f(S^l (D^l{}\Transpose \pmb{k}^l))), l=1,\ldots,L,\\
&       \mathbf{c}(\mathbf{s}, \mathbf{F}) = W_k^{L+1} \pmb{k}^{L+1} + \pmb{b}_k^{L+1}, 
    \end{split}
\end{equation}
According to equation~(\ref{aaa})~(\ref{bbb}), the $\mathbf{\dot{s}_0}$ and $\mathbf{c}$ can be calculated simultaneously at once after other factors are settled, such as the width of each layer and rank $r$, etc. 

Overall, our architecture is similar to the FC neural network. However, with the same width and depth, due to the introduced $\textsc{svd-FC}^{l}$ and $\textsc{sd-FC}^{l}$ blocks, the parameters in our architecture can be much less than the FC neural network, which makes ours not as redundant as FC neural network. Furthermore, using SVD techniques in our architecture brings noise immunity benefits \cite{golub2013matrix}. These two advantages are extensively validated in our experiments. 

\subsection{Two-phase training algorithm}\label{sec:train_alg}
Except for our neural network architecture, we also propose a two-stage training algorithm to adapt to it.
Phase 1 aims to learn the SVD matrixes $S^l, D^l$, where $l=1,\ldots, L$, and orthogonal relation between $\mathbf{\dot{s}_0}$ and $\mathbf{c}$, i.e, equation~(\ref{eq:constant-of-motion}). Phase 2 is for training these two networks to solve the whole dynamical system. 
Table~\ref{tab:param_summary} lists the sets of model parameters that are being trained in each phase.

In Phase 1, we train all the parameters of these two neural networks. Since we force to apply the SVD constraint for the hidden layers, the matrix $S^l$ and $D^l$ should satisfy the following semi-orthogonal constraint, $S^l{}\Transpose S^l = I$ and $D^l{}\Transpose D^l = I$, where $I$ is an $r \times r$ identity matrix.
Therefore, the loss function, denoted as $\mathcal{L}_1$,  in Phase 1 is as follows:
\begin{equation}\label{eq:L1}
\begin{split}
\mathcal{L}_1=&\sum_{i=1}^{n_c} \left\lVert\nabla c_i(\mathbf{s} + \mathbf{\Tilde{s}}) \cdot \mathbf{\dot{s}_0}(\mathbf{s} + \mathbf{\Tilde{s}})\right\rVert^2 + \\ 
&\sum_{l=1}^{L} (w_1 \| S^l{}\Transpose S^l - I \|_F^2  + w_2 \| D^l{}\Transpose D^l - I \|_F^2),
\end{split}
\end{equation}
where subscript $F$ means Frobenius norm and $w_1= w_2 = \frac{1}{r^2}$. 

In Phase 2, we continue training these two networks for approximating the solutions of the dynamical system. After Phase 1 training, we fix the $S^l$ and $D^l$ in hidden layers, and reinitialize other parameters in the neural networks to avoid local minimum, shown in table~\ref{tab:param_summary}. The loss function, labeled as $\mathcal{L}_2$, in Phase 2 is expressed as follows:
\begin{equation}\label{eq:L2}
\mathcal{L}_2 = \left\lVert\mathbf{\dot{s}} -  \mathbf{\hat{\dot{s}}}\right\rVert^2 + w_0 \left\lVert\mathbf{\dot{s}_0} - \mathbf{\hat{\dot{s}}}\right\rVert^2.
\end{equation}
where $w_0 = 1.0$.
It does not include the orthogonality constraint between $\mathbf{\dot{s}_0}$  and  $\mathbf{c}$ in $\mathcal{L}_2$ unless otherwise stated. 
We hope the learned orthogonality constraint could guide the neural network training in Phase 2. Although our training strategy is completely different from the COMET method, it performs as well as or even better than COMET on many dynamical systems.
\begin{table}[t]
\centering
\small
\caption{Learnable parameters in each phase}\label{tab:param_summary}
\begin{tabular}{l|l}
\toprule
Phase 1 & $\{(S^l, V^l, D^l)\}_{l=1}^{L}$, $W_{h,k}^0, W_{h,k}^{L+1}, \pmb{b}_{h,k}^0, \pmb{b}_{h,k}^{L+1}$ \\
\midrule
Phase 2 & $\{V^l\}_{l=1}^L$, $W_{h,k}^0, W_{h,k}^{L+1}, \pmb{b}_{h,k}^0, \pmb{b}_{h,k}^{L+1}$ \\
\bottomrule
\end{tabular}
\end{table}

\section{Experiments}
\subsection{Experimental setup}
\label{sec:experiment-setup}
We conducted experiments from four aspects on various systems to compare our approach with COMET comprehensively. The details are as follows:
\begin{itemize}
\item \textbf{Toy cases} Six different simple and classical dynamical systems were selected in this case:
(1) mass-spring, (2) 2d pendulum, (3) damped pendulum, (4) two body, (5) nonlinear spring 2d, and (6) Lotka-Volterra. The description for each system and the neural network utilized for the COMET method can be found in the appendix of the COMET paper \cite{kasim2022constants}.

\item \textbf{Systems with external influences} 
Unlike toy cases, the input of the neural networks is the concatenation of the state $\mathbf{s}$ and the external force $\mathbf{F}$. The chosen system is a 2d pendulum with periodic time-dependent external force. 

\item \textbf{Finding the number of constants of motion} 
We tested the capability of our method to find the number of constants of motion on three systems, namely damped pendulum, two body, and nonlinear spring 2d. 

\item \textbf{More complex cases} To validate our method more broadly, we conducted experiments on two complex cases: Korteweg-De Vries (KdV) equation and learning from pixels. 

\end{itemize}

For the first two cases, the training data were generated by simulating the system's dynamics from t = 0 to t = 10. 
During the simulations, the training data, i.e. the states $\mathbf{s}$ and the states change rate $\mathbf{\hat{\dot{s}}}$, were calculated analytically and were added Gaussian noise, $\mathcal{N}(\mathbf{0}, \sigma^2 \mathbf{I})$, with standard deviation (std) $\sigma=0.05, 0.1, 0.2$. On the contrary, we set $\sigma = 0.0$ in the third case to mimic the true evolution sufficiently. For more complex cases, we will explain the training data in their corresponding subsections. All the experiments are implemented using PyTorch on a remote server with NVIDIA A100 GPU (40GB memory) and Intel Xeon Gold 5320 (26 cores). The Pytorch's built-in Adam optimizer and a learning rate with a cosine annealing scheduling strategy are adopted to train our model. The maximum learning rate is $1 \times 10^{-3}$ and the minimum value of the learning rate is $1 \times 10^{-5}$. We set 2 hidden layers with 250 neurons for our method. Different ranks $r$ are chosen for different systems. SiLU activation function is utilized across all the experiments for Meta-COMET. COMET is set as the baseline model, and other mentioned advanced methods after COMET, such as FINDE  and ConCerNet, are not currently compared since this paper mainly focuses on improving COMET. We will make more comparisons in our upcoming paper. Our implementation and experiments are all based on the COMET's public code.

\subsection{Experimental results}
\subsubsection{Toy cases}
For each system, we run another 100 simulations from $t=0$ to $t=100$, which is 10 times longer than the training, to test the performance of our method. 1000 sample points are collected from each simulation under random initial conditions by setting different seeds. The root mean squared errors (RMSE) of the state predictions for each system are summarized in table \ref{tab:mse-results1}. In addition, we also calculate the number of parameters of each model which is independent of noise, shown in table \ref{tab:mse-results2}. 

\begin{table*}[t]
\centering
\footnotesize
\vspace{-2mm}
\begin{center}
\setlength{\arrayrulewidth}{1pt}
\begin{tabular}{|c|cc|cc|cc|}
\hline
\multirow{2}{*}{Case}& \multicolumn{2}{c|}{$\sigma = 0.05$} & \multicolumn{2}{c|}{$\sigma = 0.1$} & \multicolumn{2}{c|}{$\sigma = 0.2$} \\ 
\cline{2-7}
 & COMET & Meta-COMET  & COMET & Meta-COMET  & COMET & Meta-COMET  \\ 
\hline
mass-spring  & $ 0.10^{+0.15}_{-0.09}$ & $\mathbf{0.075^{+0.01}_{-0.07}}$ & $ \mathbf{0.15^{+0.23}_{-0.13}}$  & $ 0.11^{+0.31}_{-0.09}$ & $0.10^{+0.18}_{-0.09}$ & $\mathbf{0.08^{+0.01}_{-0.03}}$  \\ [0.5ex]
\hline
2d pendulum  & $ 0.18^{+0.17}_{-0.14}$ & $ \mathbf{0.037^{+0.10}_{-0.03}}$ & $ 0.17^{+0.38}_{-0.15}$  & $ \mathbf{0.045^{+0.19}_{-0.04}}$ & $0.31^{+0.33}_{-0.27}$ & $\mathbf{0.067^{+0.38}_{-0.05}}$  \\ [0.5ex]
\hline
damped pendulum & $ \mathbf{0.007^{+0.014}_{-0.005}}$ & $ 0.0036^{+0.019}_{-0.0025}$ & $ 0.018^{+0.045}_{-0.008}$  & $ \mathbf{0.007^{+0.038}_{-0.005}}$ & $0.037^{+0.54}_{-0.026}$ & $\mathbf{0.015^{+0.06}_{-0.01}}$  \\ [0.5ex]
\hline
two body  & $ \mathbf{0.42^{+0.48}_{-0.39}}$ & $ 0.43^{+0.57}_{-0.32}$ & $ 0.65^{+0.28}_{-0.55}$  & $ \mathbf{0.58^{+0.34}_{-0.50}}$ & $0.65^{+0.42}_{-0.58}$ & $\mathbf{0.63^{+0.43}_{-0.57}}$  \\ [0.5ex]
\hline
nonlinear spring 2d & $ \mathbf{0.23^{+0.40}_{-0.18}}$ & $0.52^{+0.61}_{-0.46}$ & $ \mathbf{0.69^{+0.44}_{-0.38}}$  & $ 0.62^{+0.58}_{-0.51}$ & $0.75^{+0.43}_{-0.32}$ & $\mathbf{0.73^{+0.43}_{-0.50}}$  \\ [0.5ex]
\hline
Lotka-Volterra  & $ \mathbf{0.048^{+0.055}_{-0.041}}$ & $0.044^{+0.08}_{-0.017}$ & $ 0.038^{+0.11}_{-0.03}$  & $ \mathbf{0.09^{+0.03}_{-0.06}}$ & $0.12^{+0.26}_{-0.11}$ & $\mathbf{0.11^{+0.06}_{-0.09}}$  \\ [0.5ex]
\hline
\end{tabular}
\end{center}
\caption{RMSE of 100 randomly initialized simulations for each system and each method with different $\sigma$. The median values and their 95\% percentiles are depicted, i.e. lower and upper bounds are 2.5\% and 97.5\% percentiles, respectively. The better upper bounds between these two methods are highlighted with bold values. The values for COMET with $\sigma=0.05$ are borrowed from its original paper.}
\label{tab:mse-results1}
\end{table*}

\begin{table}[t]
\centering
\footnotesize
\begin{center}
\setlength{\arrayrulewidth}{1pt}
\begin{tabular}{|c|cc|}
\hline
\multirow{2}{*}{Case} & \multicolumn{2}{c|}{Number of parameters}  \\ 
\cline{2-3}
 & COMET & Meta-COMET    \\ 
\hline
mass-spring &  189753 & 12273 ($r=10$) \\ 
\hline
2d pendulum  &  191257 & 14277 ($r=10$) \\ 
\hline
damped pendulum   &  191006 & 14026 ($r=10$)  \\ 
\hline
two body  & 194265 & 28305 ($r=20$) \\ 
\hline
nonlinear spring 2d  & 191006 & 34066 ($r=30$)   \\ 
\hline
Lotka-Volterra  &  189753 & 12273 ($r=10$) \\ 
\hline
\end{tabular}
\end{center}
\caption{The total number of parameters (and rank $r$) for each case and each method.}
\label{tab:mse-results2}
\end{table}

\begin{figure}[t]
    \begin{subfigure}{\linewidth}
        \centering
 \parbox{\textwidth} { \hspace{0.1cm} {\tiny COMET}  \hspace{1.4cm}  {\tiny Meta-COMET} \hspace{1.4cm} {\tiny True}} \\
        \includegraphics[width=\linewidth]{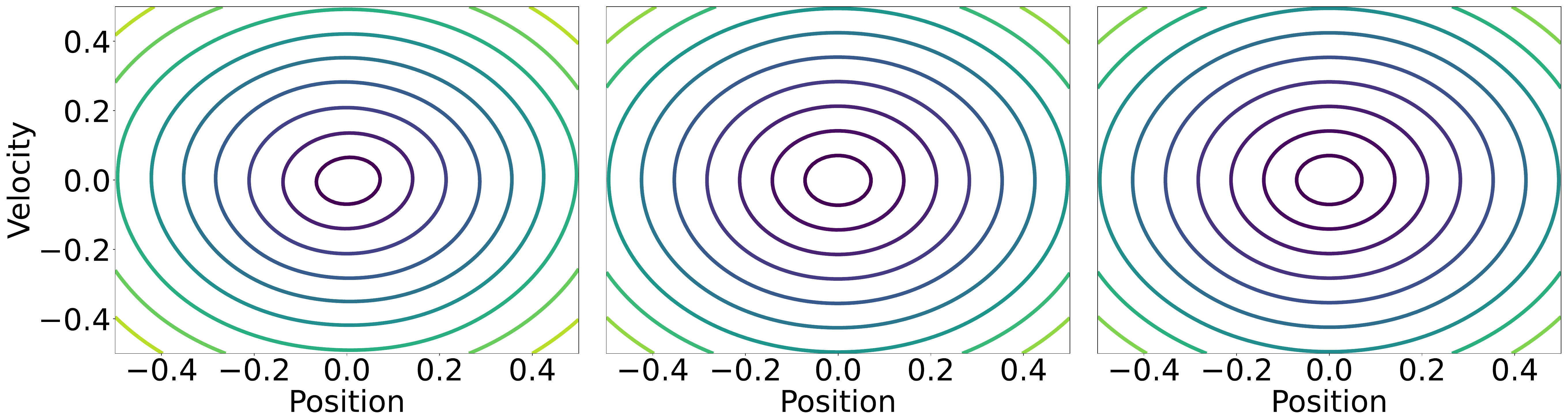}
    \end{subfigure}
    \begin{subfigure}{\linewidth}
        \flushleft
        \includegraphics[width=0.686\linewidth]{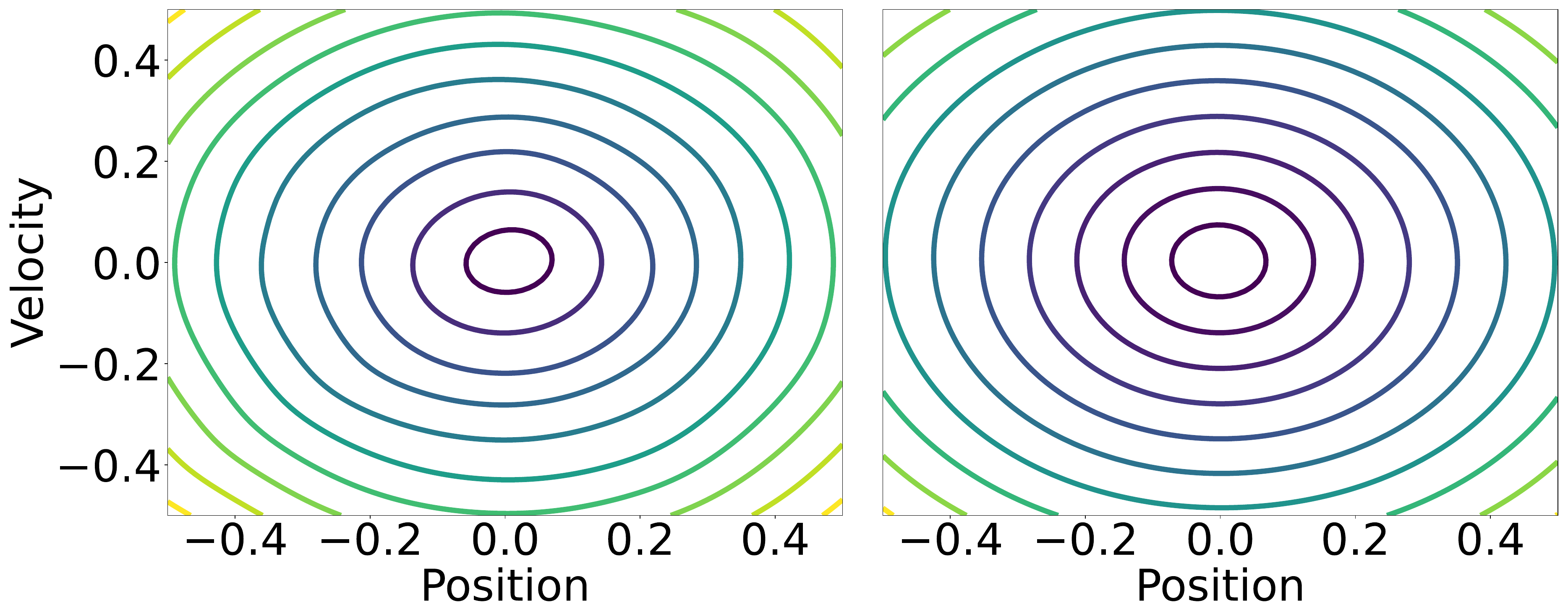}
    \end{subfigure}
        \begin{subfigure}{\linewidth}
       \flushleft
        \includegraphics[width=0.686\linewidth]{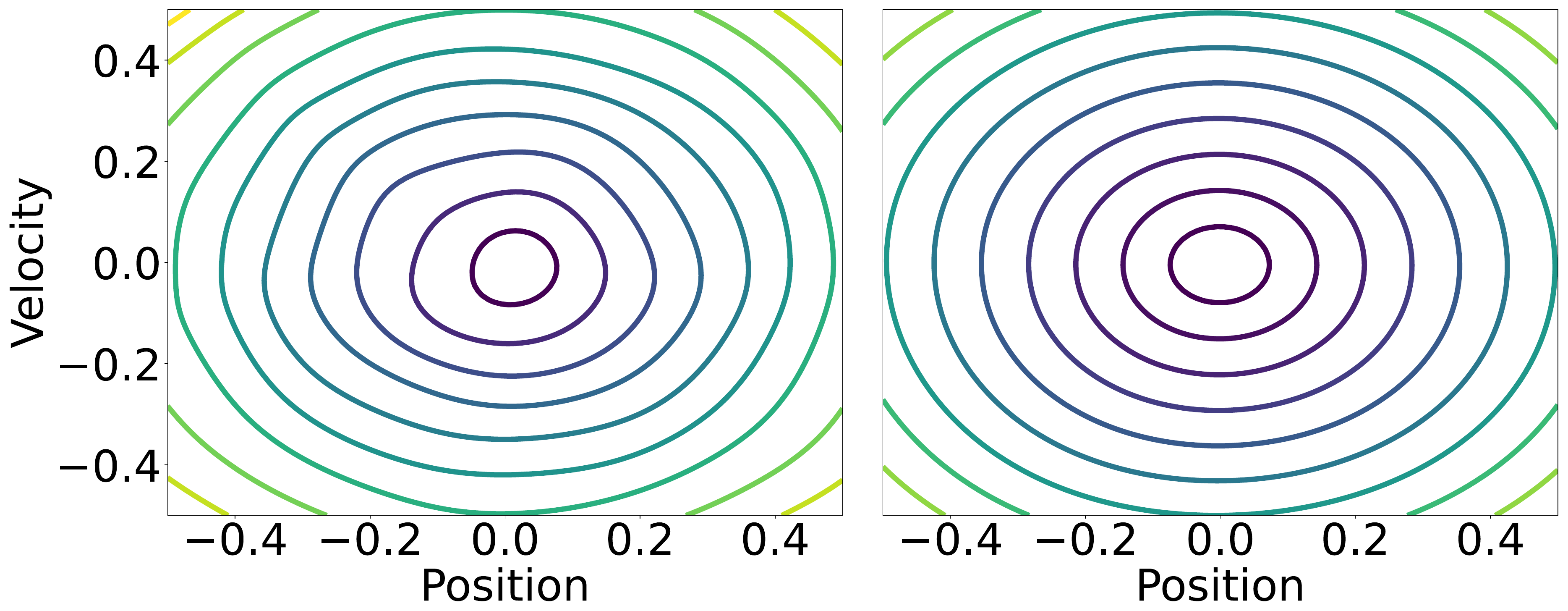}
    \end{subfigure}
    \caption{With different $\sigma$ (equals to 0.05, 0.1, 0.2 from top to bottom), the contour plot of constant of motion discovered by Meta-COMET and COMET for the mass-spring case. The left column is for COMET. The middle column is for Meta-COMET. The last column is the ground truth.}
    \label{fig:discovered-com1}
\end{figure}
\begin{figure}[t]
    \begin{subfigure}{\linewidth}
        \centering
 \parbox{\textwidth} { \hspace{0.1cm} {\tiny COMET}  \hspace{1.4cm}  {\tiny Meta-COMET} \hspace{1.4cm} {\tiny True}} \\
        \includegraphics[width=\linewidth]{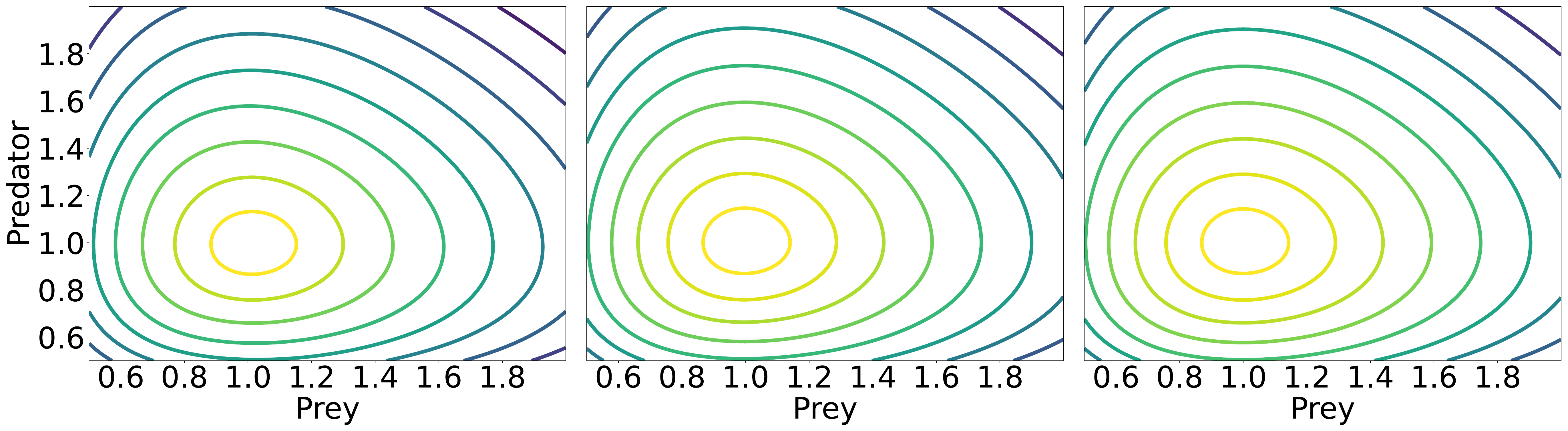}
    \end{subfigure}
    \begin{subfigure}{\linewidth}
        \flushleft
        \includegraphics[width=0.686\linewidth]{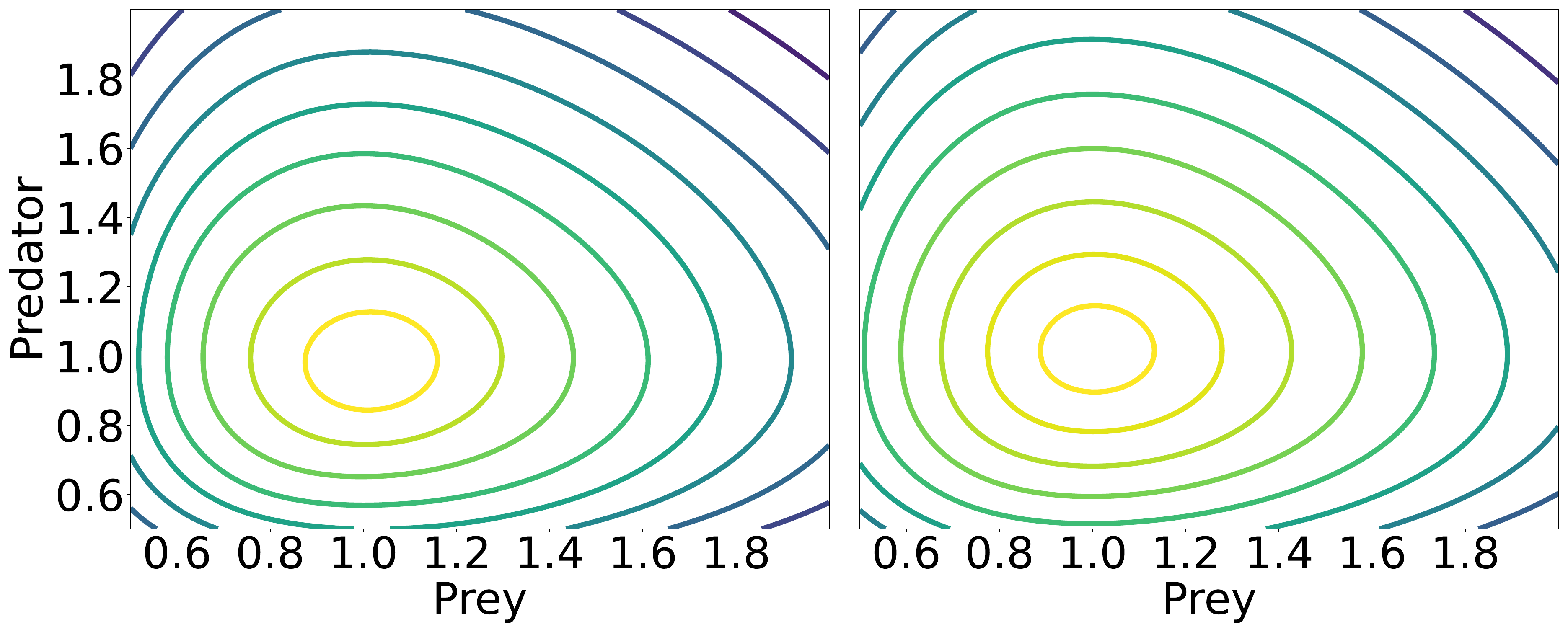}
    \end{subfigure}
        \begin{subfigure}{\linewidth}
        \flushleft
        \includegraphics[width=0.686\linewidth]{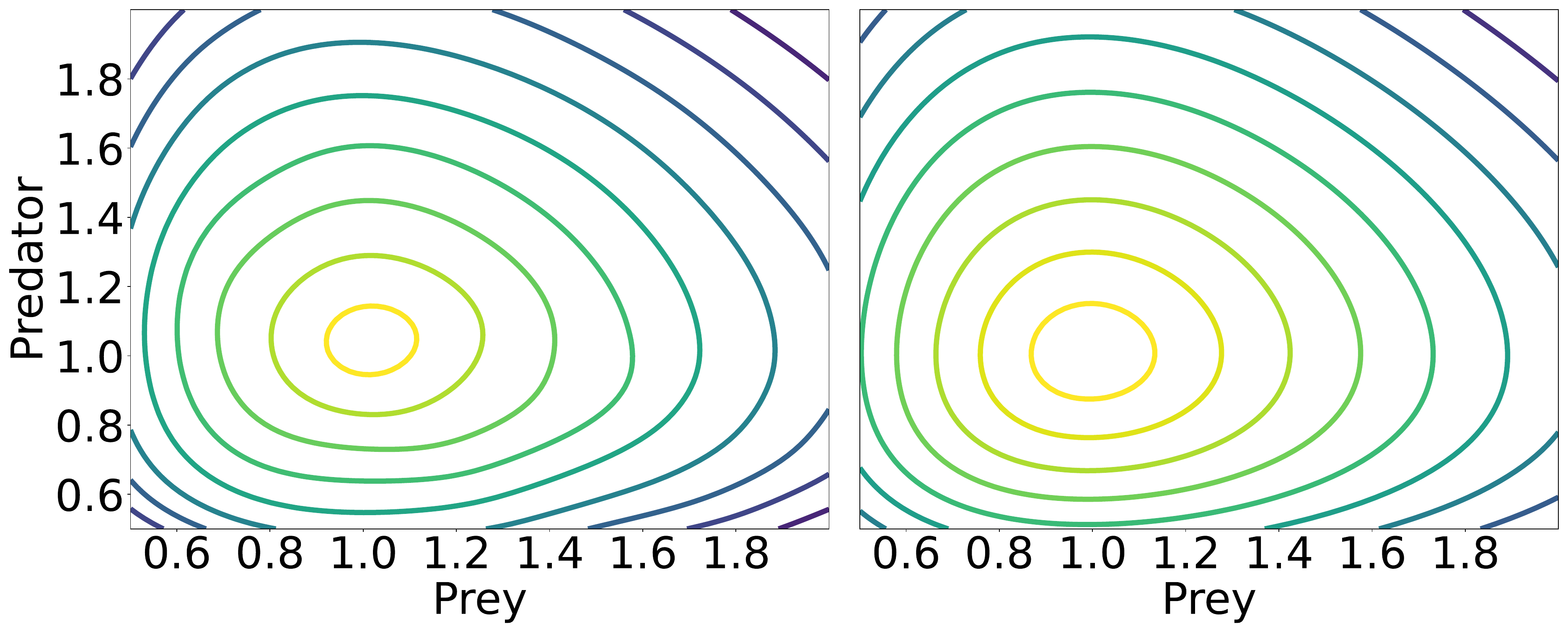}
    \end{subfigure}
    \caption{Same as Figure~\ref{fig:discovered-com1} except for describing the Lotka-Volterra case.}
    \label{fig:discovered-com2}
\end{figure}
compared to the COMET method, our method 
achieves better prediction across all the test cases when $\sigma=0.2$. Meanwhile, most of our predicted results are superior to those of COMET for $\sigma = 0.1$. 
These results show that our scheme not only inherits the advantage of COMET, but also outperforms it in terms of noise robustness. Furthermore, the parameters of the neural network used in Meta-COMET are much less than COMET according to the table~\ref{tab:mse-results2}. For example, the parameters are reduced by approximately 14 times for the mass-spring and Lotka-Volterra cases. Even in the case of the nonlinear spring 2d, which has the largest number of parameters, 34066, among these systems using the Meta-COMET method, the parameters used are still only about 18 percent of those of COMET. 
\begin{figure*}
    \centering
    \begin{subfigure}{0.25\textwidth}
        \centering
        \parbox{\textwidth} {\hspace{1.0cm} {\tiny Energy (mass-spring)}} \\
        \includegraphics[width=\linewidth]{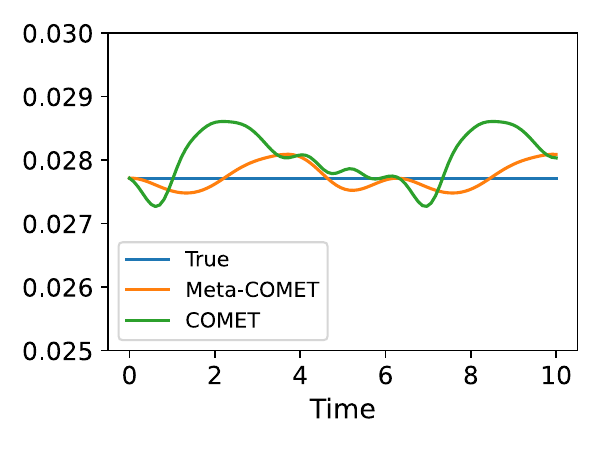}
        \includegraphics[width=\linewidth]{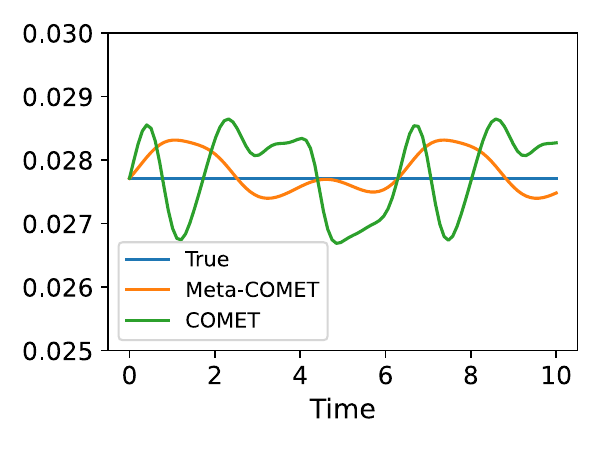}
        \includegraphics[width=\linewidth]{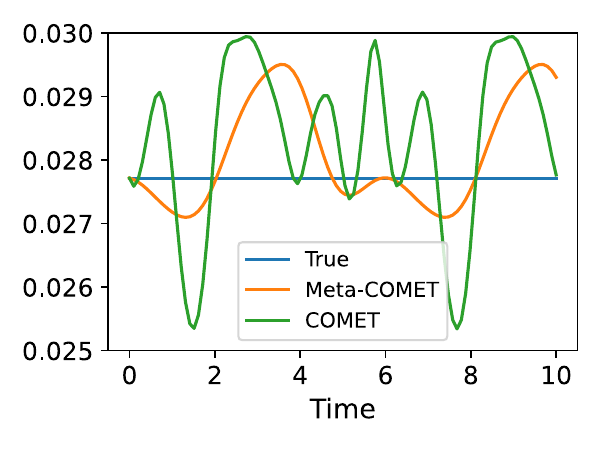}
        \caption{}
    \end{subfigure}%
    \begin{subfigure}{0.75\textwidth}
        \centering
        \parbox{\textwidth} {\hspace{1.0cm} {\tiny Energy (2d-pendulum)}\hspace{2.cm} {\tiny Length (2d-pendulum)}\hspace{2.0cm} {\tiny Velocity angle (2d-pendulum)}} \\
        \includegraphics[width=\linewidth]{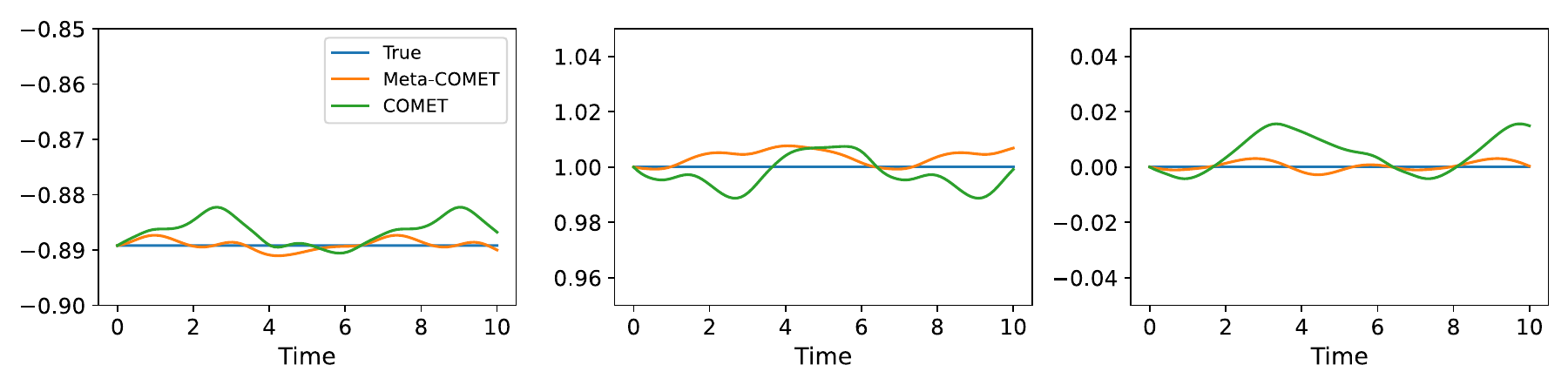}
        \includegraphics[width=\linewidth]{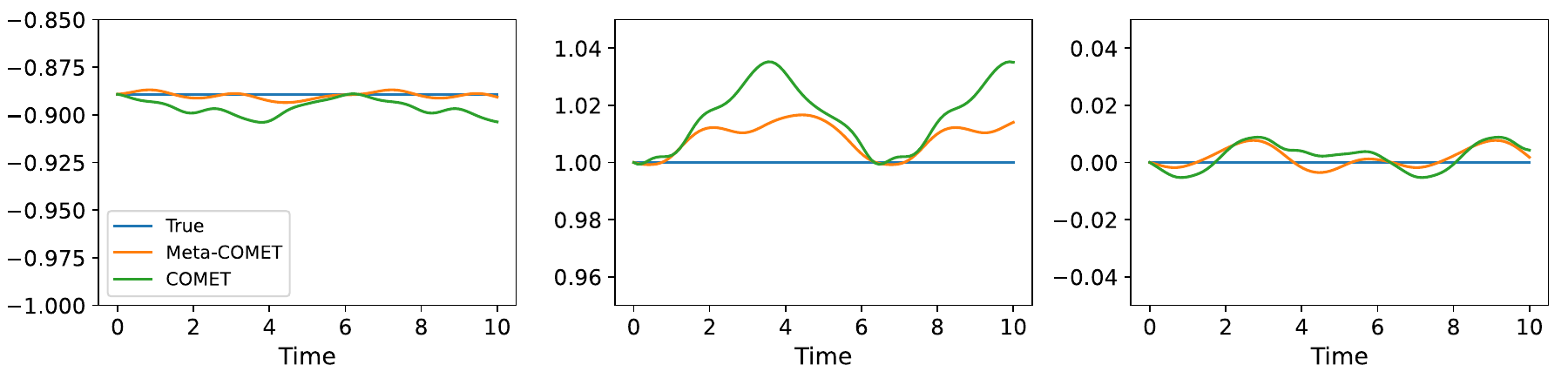}
        \includegraphics[width=\linewidth]{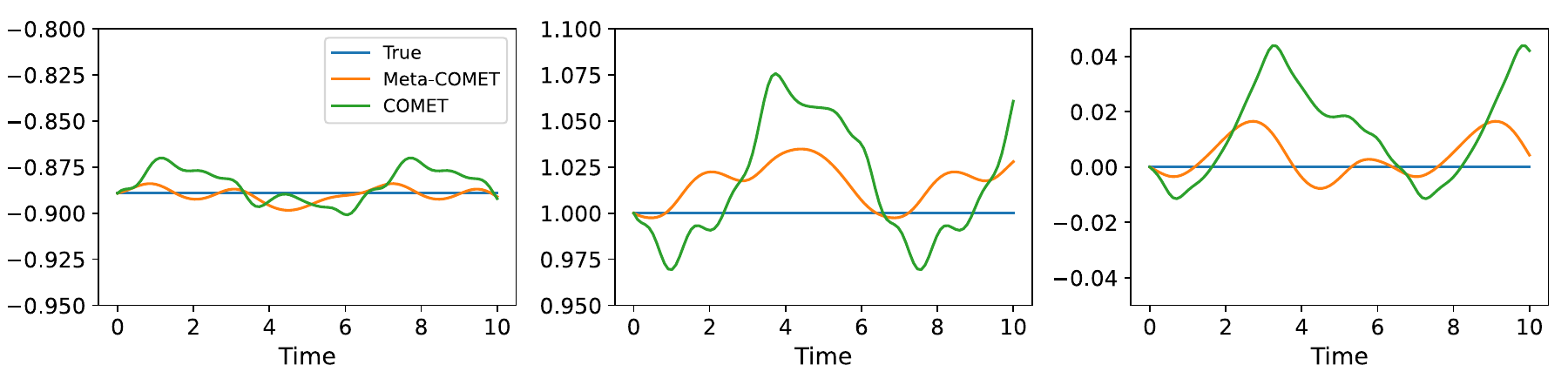}
        \caption{}
    \end{subfigure}
    \begin{subfigure}{\textwidth}
        \centering
         \parbox{\textwidth} {\hspace{1.cm} {\tiny Energy (two-body)}\hspace{2.cm} {\tiny x-momentum (two-body)}\hspace{2.cm} {\tiny y-momentum  (two-body)} \hspace{1.5cm} {\tiny Angular momentum  (two-body)}} \\
        \includegraphics[width=\linewidth]{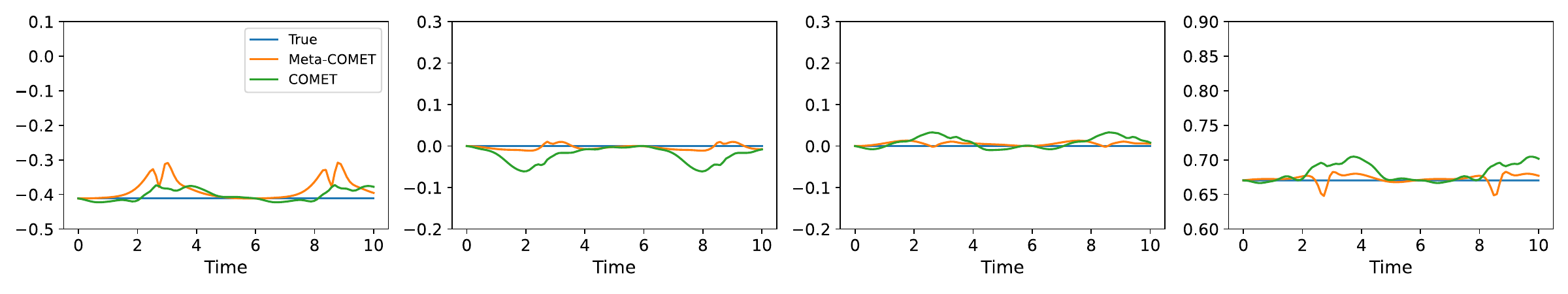}
         \includegraphics[width=\linewidth]{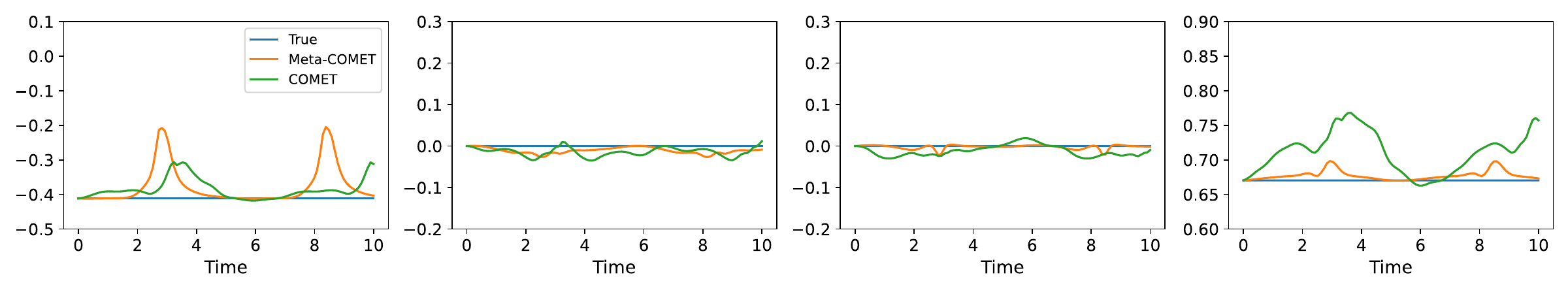}
          \includegraphics[width=\linewidth]{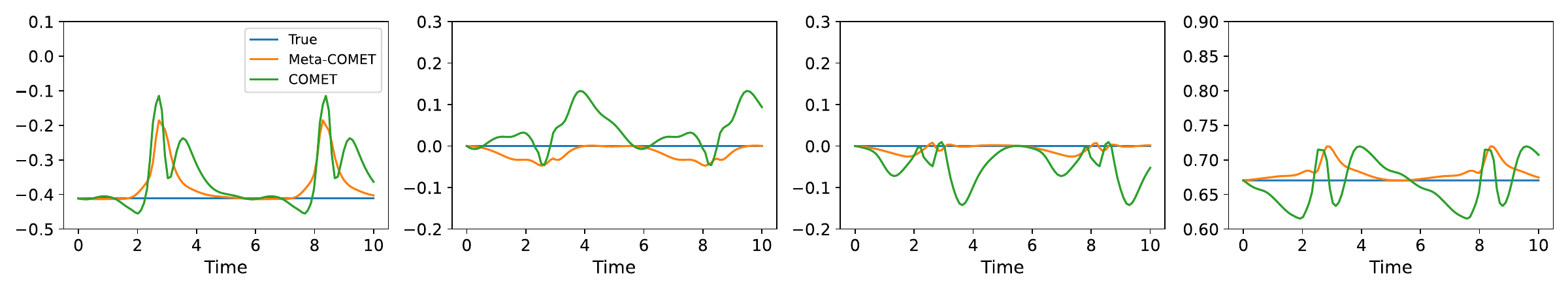}
        \caption{}
    \end{subfigure}
    \caption{The evolution of constants of motion calculated by  Meta-COMET and COMET for (a) mass-spring, (b) 2d pendulum, and (c) two body system. $\sigma$ is equal to 0.05, 0.1, and 0.2 from top to bottom in each case.}
    \label{fig:constants-of-motion}
\end{figure*}

Figure~\ref{fig:discovered-com1} and~\ref{fig:discovered-com2} show the discovered constant of motion in the
mass-spring and Lotka-Volterra cases. From these Figures, similar to the COMET, Meta-COMET can successfully discover the constants of motion from the data. In addition, with the increased noise in the training data, the curves of Meta-COMET remain similar to the ground truth, however, the corresponding curves for COMET begin to distort and deviate from the true ones in both cases. Since these curves represent the intrinsic mathematical relation between the systems' states, it indicates that our method can still capture the subtle relation between these states even with a larger introduced noise, i.e. $\sigma=0.2$. On the contrary, the COMET method begins to misunderstand these specific correlations.

\begin{figure}[t]
    \centering
        \begin{subfigure}{\linewidth}
        \centering
 \parbox{\textwidth} { \hspace{0.1cm} {\tiny COMET}  \hspace{1.4cm}  {\tiny Meta-COMET} \hspace{1.4cm} {\tiny True}} \\
        \includegraphics[width=\linewidth]{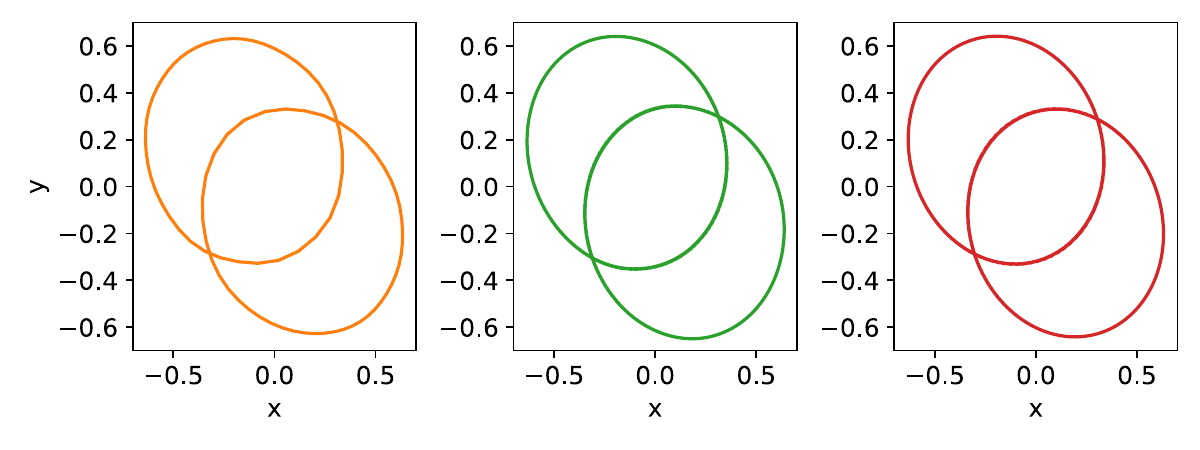}
    \end{subfigure}
    \begin{subfigure}{\linewidth}
        \flushleft
        \includegraphics[width=0.686\linewidth]{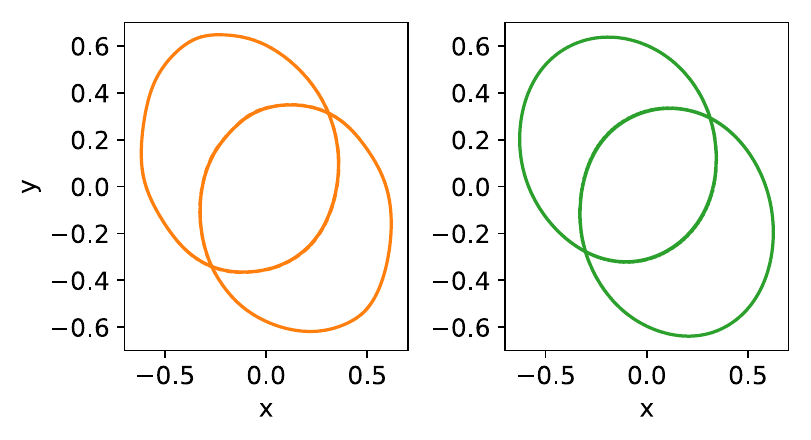}
    \end{subfigure}
        \begin{subfigure}{\linewidth}
        \flushleft
        \includegraphics[width=0.686\linewidth]{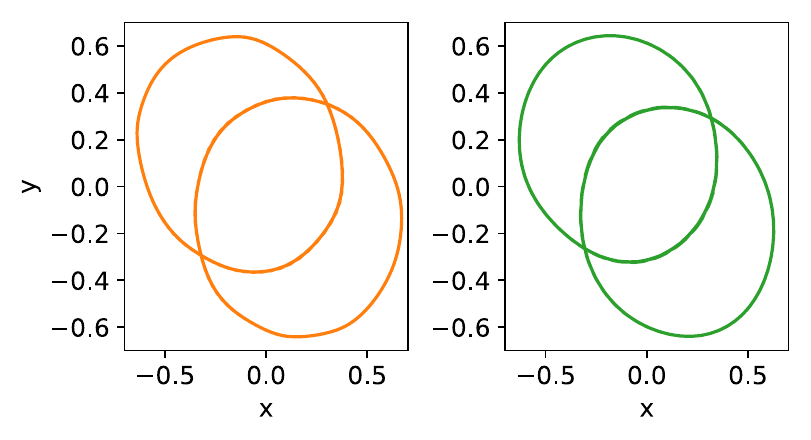}
    \end{subfigure}
    \caption{From the left column to the right column, it represents the motion trajectory of the simulated two body system from $t=0$ to $t=20$ predicted by the COMET, Meta-COMET and its analytical equation when $\sigma = 0.05, 0.1, 0.2$ (from top to bottom).}
    \label{fig:two-body-trajectory}
\end{figure}

Figure~\ref{fig:constants-of-motion} shows the evolution of the known constants of motion for each method in the mass-spring, 2d pendulum, and the two body cases. Due to the added noise in the training data and symmetries of the systems, the periodic variation from the true constants of motion can be observed. In each case, $\sigma$ is equal to 0.05, 0.1, 0.2 from top to bottom. Consistent with intuition, these curves deviate more from the true constant values as the noise increases. 

In the mass-spring case, there is only one constant of motion. As shown in Figure~\ref{fig:constants-of-motion}(a), the Meta-COMET outperforms the COMET in each case when $\sigma = 0.05, 0.1, 0.2$.
A similar story can be found in Figure~\ref{fig:constants-of-motion}(b) where it shows three constants of motion for the 2d pendulum case. It illustrates that the predicted results from Meta-COMET have smaller deviations from the true conserved values in all these three constants of motion. More specifically, for example,
the maximum deviation in COMET values is approximately more than twice that of Meta-COMET when predicting velocity angles with $\sigma = 0.2$. 
In Figure~\ref{fig:constants-of-motion}(c), except for the Energy value in the two body cases, Meta-COMET does better across all the other cases compared to the COMET. 
What is even more exciting is that our method achieves better results in the Energy value when $\sigma = 0.2$. This phenomenon highlights the advantage of noise robustness of our method. In addition, with the increasing noise, our method still captures the features of x-momentum and y-momentum very well, while these two values predicted by the COMET method have a larger increase when $\sigma = 0.1$ changes to $\sigma = 0.2$.

In Figure~\ref{fig:two-body-trajectory}, we depict the motion trajectory of the simulated two body system from t = 0 to t = 20. When $\sigma=0.05$, both methods can recover the true trajectory of the two body system. However, once $\sigma$ increases to 0.2, the COMET method obtains a relatively more irrational trajectory. In contrast, our predicted result still behaves similarly to the ground truth. It again exemplifies that Meta-COMET is more robust to the noise than COMET. 

\subsubsection{Systems with external influences}
\begin{figure*}[t] 
    \centering
    \parbox{\textwidth} {\hspace{1.2cm} {\tiny Energy (forced-2d-pendulum)}\hspace{3.cm} {\tiny Length (forced-2d-pendulum)}\hspace{3.0cm} {\tiny Velocity angle (forced-2d-pendulum)}} \\
    \includegraphics[width=\linewidth]{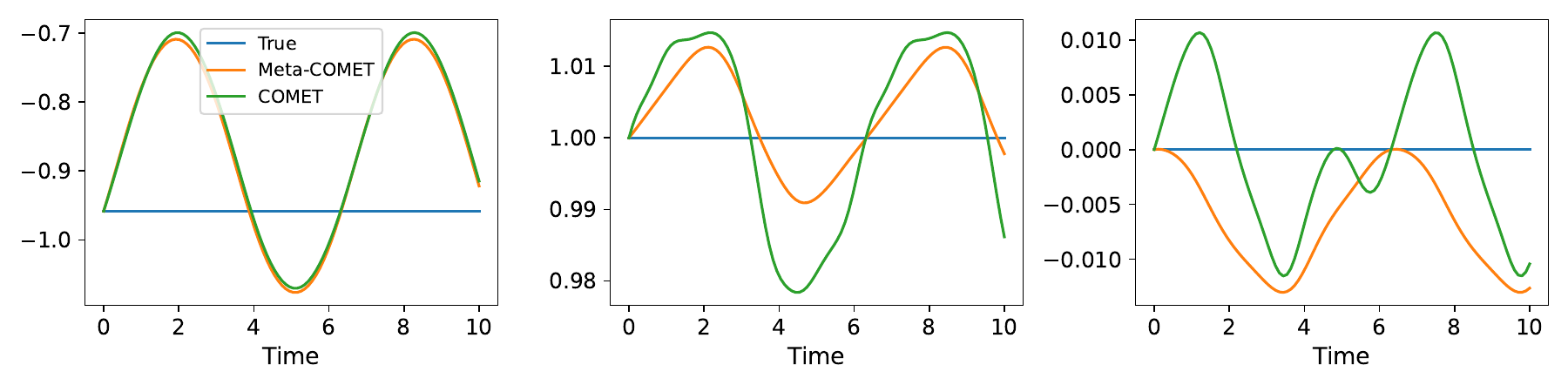}
    \includegraphics[width=\linewidth]{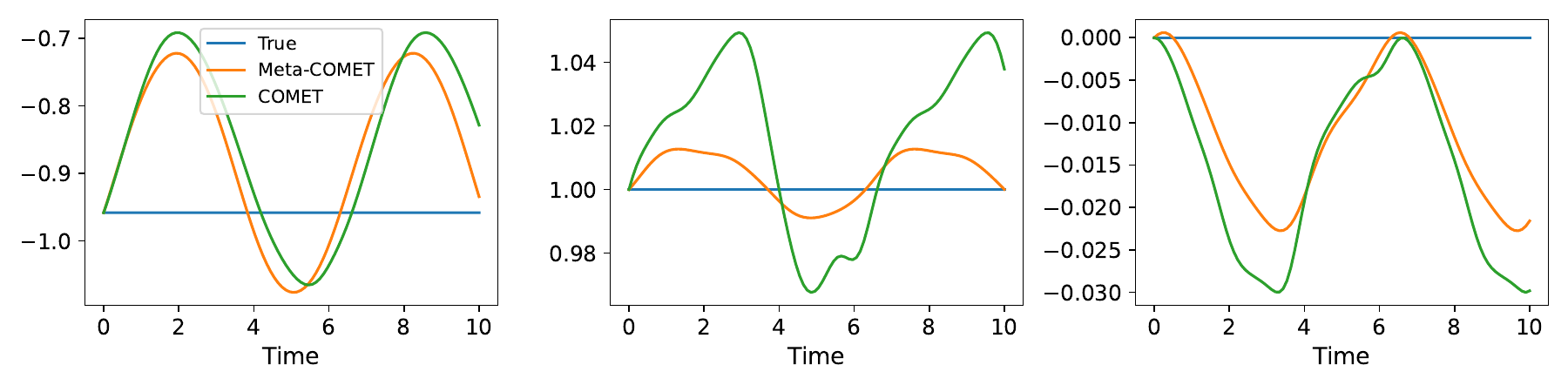}
    \includegraphics[width=\linewidth]{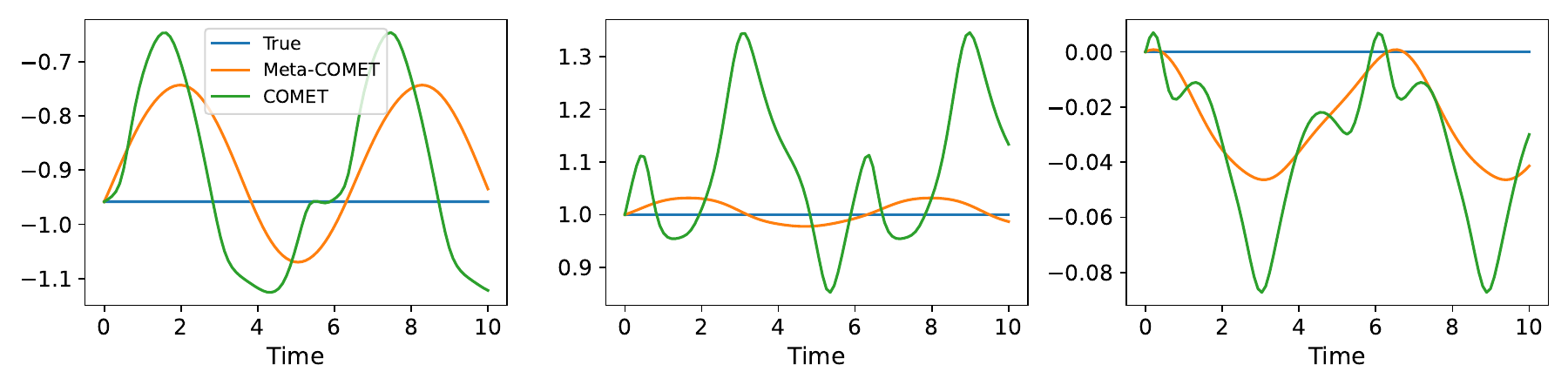}
    \caption{Constants of motion of the forced 2D pendulum case calculated using COMET and Meta-COMET with $\sigma = 0.05, 0.1, 0.2$ from top to bottom.}
    \label{fig:com-forced-system}
\end{figure*}
Just as mentioned in the methodology section, the COMET method can be extended to the system with external force by simply modifying $\mathbf{\dot{s}_0}$ and $\mathbf{c}$ to depend on the states as well as external influences.

In this case, we follow the same setting of the forced 2d pendulum in the COMET paper. The external force in the $x$-direction of 2d pendulum was $F_x(t)=a_0\cos(a_1 t + a_2)$, where $a_0$, $a_1$, and $a_2$ were random values generated from uniform distribution $\mathcal{U}(-0.5, 0.5)$, $\mathcal{U}(0, 5)$, and $\mathcal{U}(0, 2\pi)$, respectively.

Figure~\ref{fig:com-forced-system} shows the constants of motion on the test system with constant external force. As seen from Figure~\ref{fig:com-forced-system}, due to the added noise, the values of the constants of motion produced by Meta-COMET and COMET oscillate slightly around a constant offset.
Compared with COMET, Meta-COMET generally obtains a smaller error for these dynamic predictions, such as Energy value. 
Although COMET does better than our method for predicting Velocity angle when $\sigma=0.05$, its performance degrades rapidly and is surpassed by Meta-COMET after the noise increases.
Besides, when $\sigma = 0.2$, for the constant of motion, Length, the maximum error of Meta-COMET is only about 0.05. In contrast, the COMET method is approximately six times larger than that.

\subsubsection{Finding the number of constants of motion}
\label{sec:finding-ncom}
\begin{figure*}[t]
    \centering
        \parbox{\textwidth} {\centering \hspace{0.0cm} {\tiny damped-pendulum}\hspace{3.5cm} {\tiny two-body}\hspace{3.5cm} {\tiny nonlin-spring-2d}} \\
    \begin{subfigure}{0.3\textwidth}
        \centering
        \includegraphics[width=\linewidth]{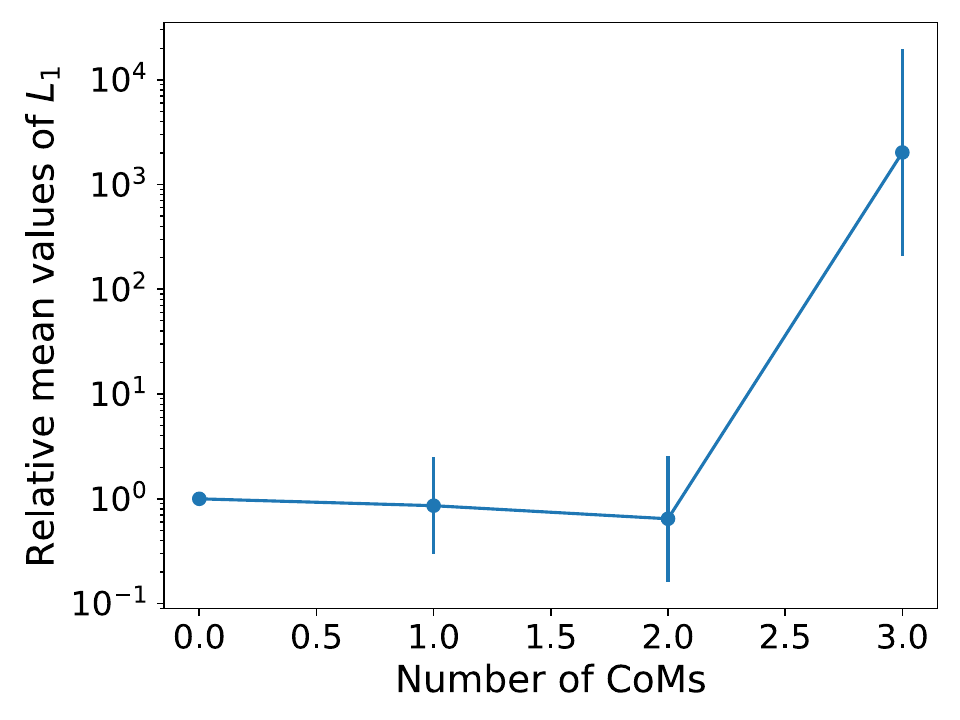}
    \end{subfigure}%
    \begin{subfigure}{0.3\textwidth}
        \centering
        \includegraphics[width=\linewidth]{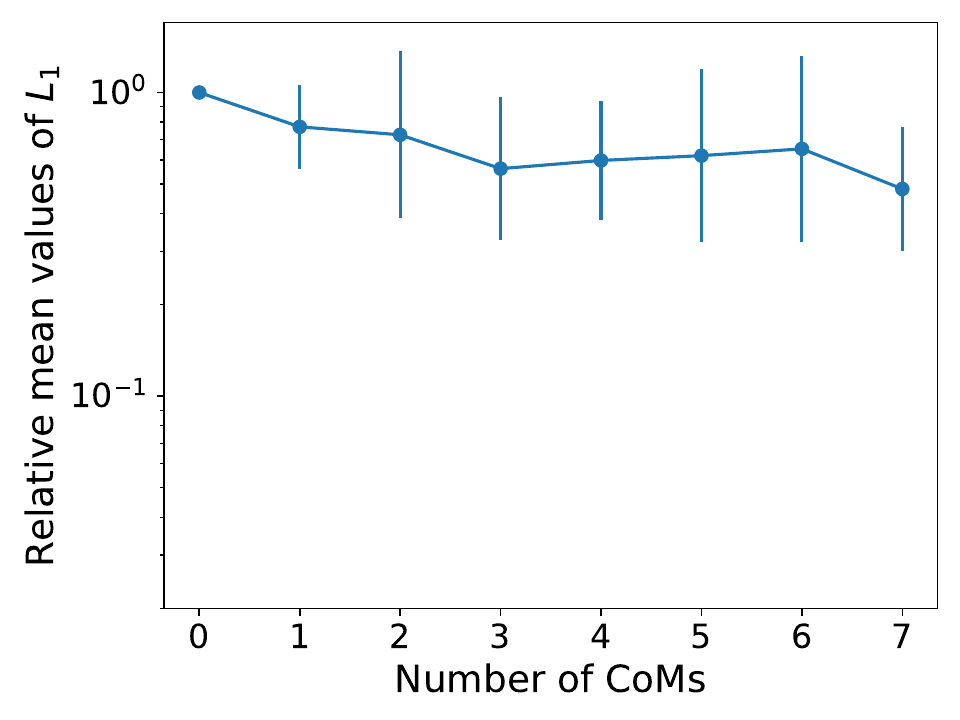}
    \end{subfigure}%
    \begin{subfigure}{0.3\textwidth}
        \centering
        \includegraphics[width=\linewidth]{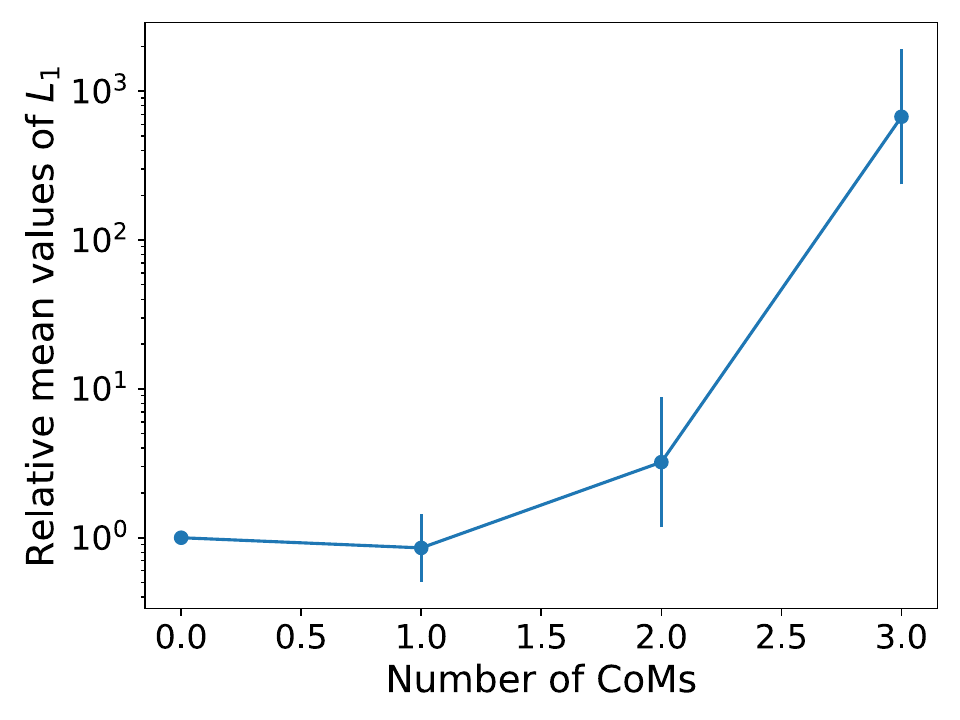}
    \end{subfigure}
    \begin{subfigure}{0.3\textwidth}
        \centering
        \includegraphics[width=\linewidth]{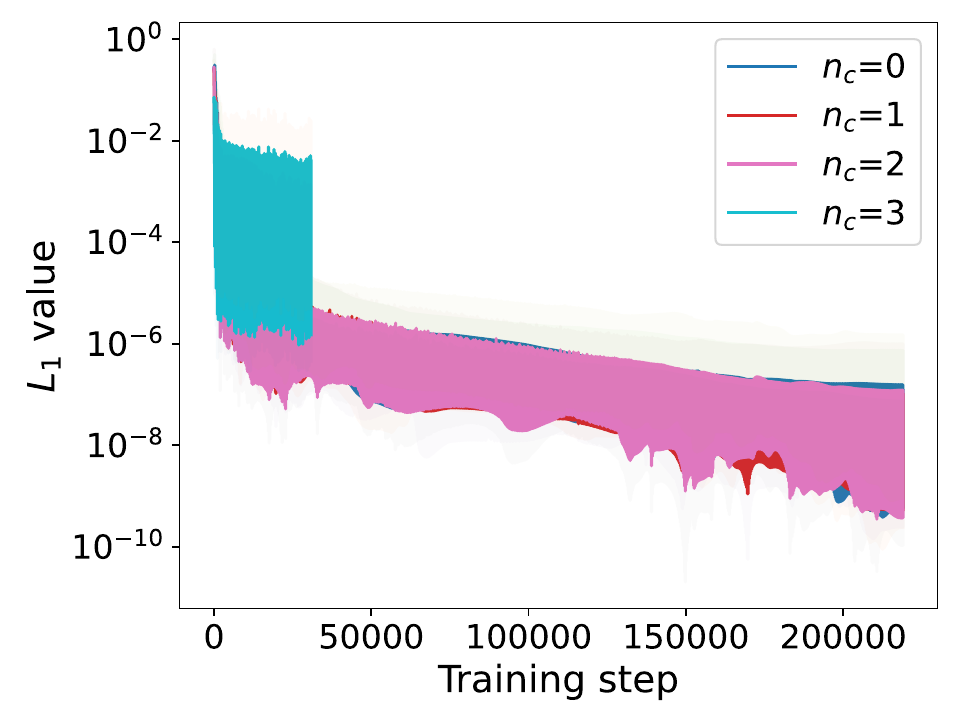}
    \end{subfigure}%
    \begin{subfigure}{0.3\textwidth}
        \centering
        \includegraphics[width=\linewidth]{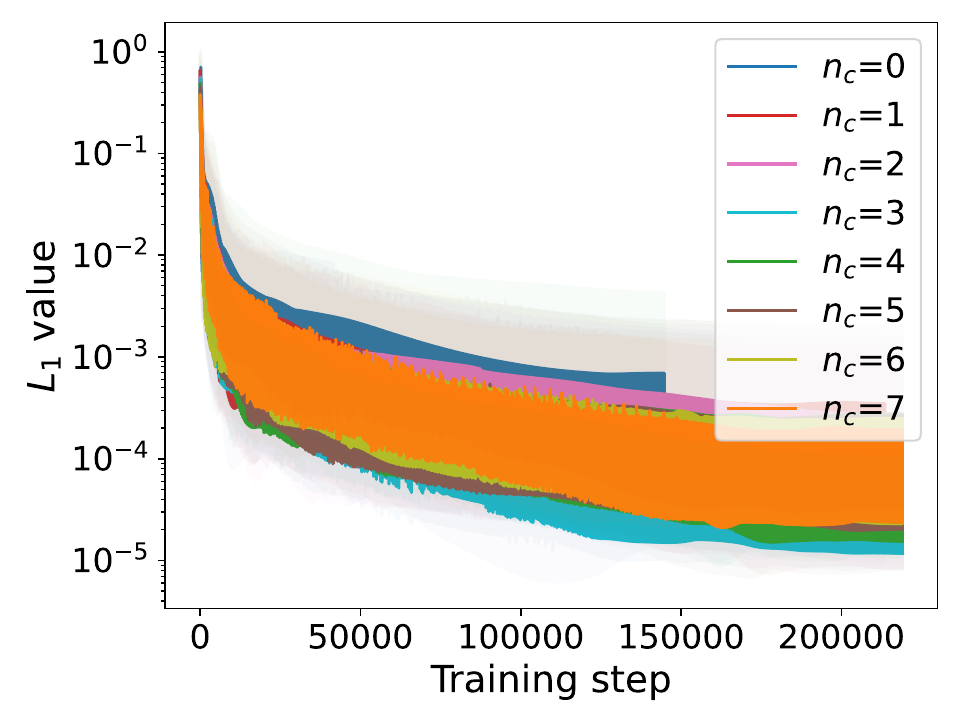}
    \end{subfigure}%
    \begin{subfigure}{0.3\textwidth}
        \centering
        \includegraphics[width=\linewidth]{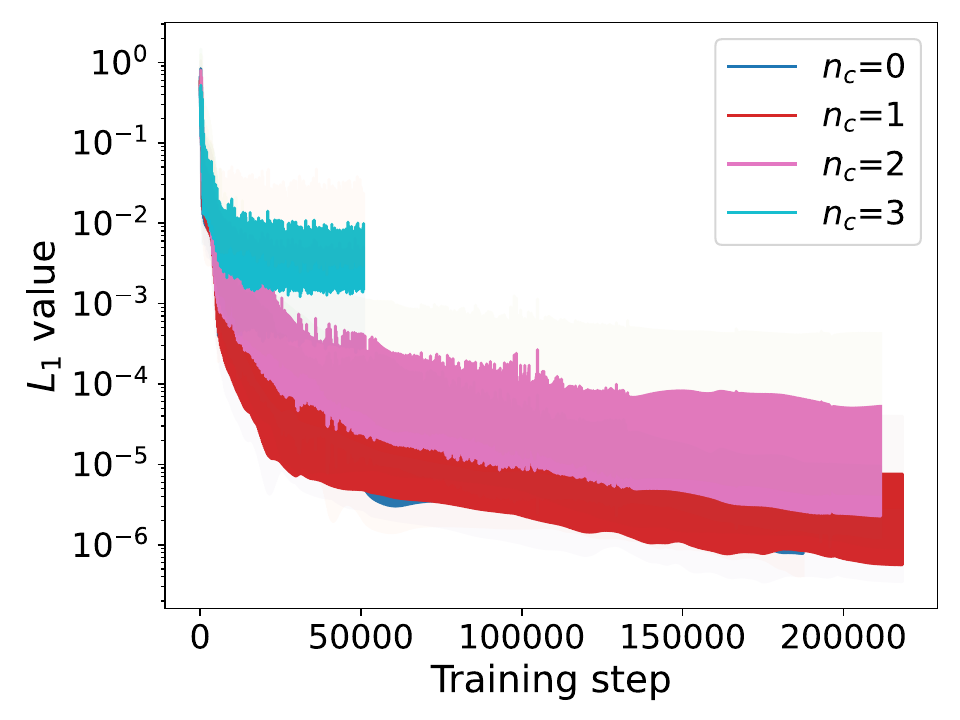}
    \end{subfigure}
    \caption{(Top row) The relative mean values of $L_1$ calculated by equation~(\ref{eq:l1}), divided by the value at $n_c=0$, for the damped pendulum, two body, and nonlinear spring 2d cases. The horizontal ordinate represents the number of constants of motion $n_c$ scanned from 0 to $n_s - 1$.
    The values and the error bars were obtained by taking the mean and std from 5 Meta-COMETs trained with different random seeds, respectively.
    (Bottom row) The values of $L_1$, averaged from 5 Meta-COMETs, during the training for various numbers of constants of motion for the damped pendulum, two body, and nonlinear spring 2d cases.
    }
    \label{fig:number-constants-of-motion}
\end{figure*}
Finding the number of independent constants of motion is an 
appealing property of the COMET. In this section, we point out that our method also has such an elegant characteristic.

Similar to COMET, we check the residual error, labeled as $L_1$, in the loss function $\mathcal{L}_2$ in Phase 2, i.e.  
\begin{align}
    \label{eq:l1}
    L_1 = \left\lVert\mathbf{\dot{s}} - \mathbf{\hat{\dot{s}}}\right\rVert^2.
\end{align}
As known from COMET, the number of constants of motion can be deduced from the sudden jump of $L_1$ values. To certify this judgment also works for Meta-COMET, we ran a few simple experiments to find the number of constants of motion for three known systems, i.e.damped pendulum, two body, and nonlinear spring 2d, which have 2, 7, and 2 constants of motion out of 4, 8, and 4 number of states. In this experiment, we train all the models for 1000 epochs in both two phases. Additionally, to speed up training, training was stopped if the validation loss did not improve during 100 epochs of consecutive training. 
All the results are visually displayed in Figure~\ref{fig:number-constants-of-motion}.

From Figure~\ref{fig:number-constants-of-motion} (top row), 
for the damped pendulum, and nonlinear spring 2d system, once 
the number of constants of motion is set above a certain number, the value of $L_1$ suddenly increases compared to the values with $n_c=0$. However, for the two body system, the value of $L_1$ remains flat since it has exactly 7 constants of motion. It indicates the actual number of constants of motion. 
A similar conclusion can also be drawn from the bottom row of Figure~\ref{fig:number-constants-of-motion}. 
When the number of constants of motion, $n_c$, was set as 3, the performance of the models for the damped pendulum, and nonlinear spring 2d system was hard to improve which resulted in a short recorded evolution of $L_1$ value. On the contrary, if $n_c$ was set to less than 3, training was inclined to continue until the end of epochs. And for the two body case, the behavior of $L_1$ values remained similar to each other when $n_c$ changed from 0 to 7. Therefore, from Figure~\ref{fig:number-constants-of-motion},  we can determine without hesitation that the numbers of motion constants for these three dynamical systems are 2, 7 and 2, respectively.

\begin{figure}[t]
    \centering
        \begin{subfigure}{\linewidth}
        \centering
        \includegraphics[width=\linewidth]{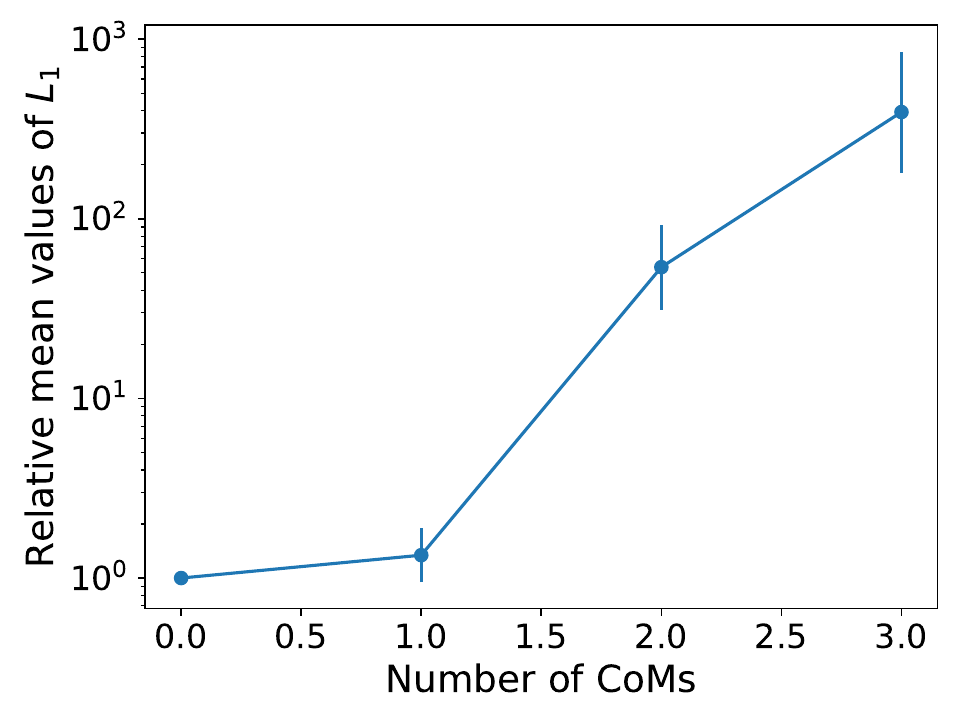}
    \end{subfigure}
    \begin{subfigure}{\linewidth}
        \flushleft
        \includegraphics[width=\linewidth]{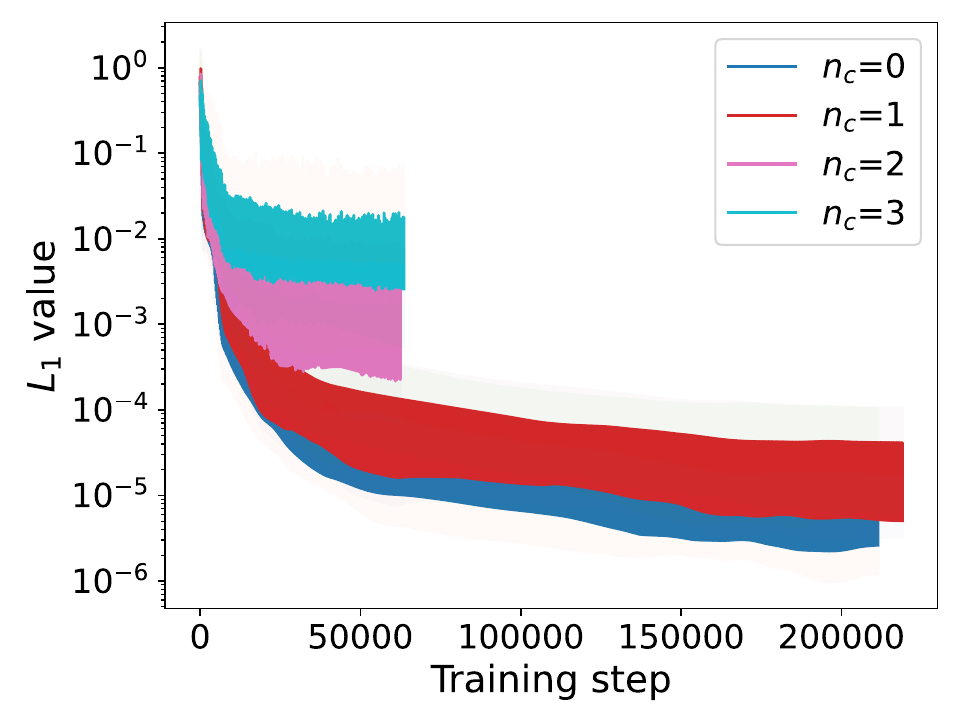}
    \end{subfigure}
    \caption{An example of the failure to find the number of constants of motion with a smaller network.}
    \label{fig:ncom-failure-mode}
\end{figure}

Besides the technique of finding the number of constants of motion, 
COMET also warns us about the limitation of this ability which is that it depends heavily on the expressive power of the neural network. 
To reproduce this phenomenon, we experimented on the nonlinear spring 2d system where the neural network only has 30 hidden neurons per layer instead of the previous 250. The results of $L_1$ values are shown in Figure~\ref{fig:ncom-failure-mode}. Unlike Figure~\ref{fig:number-constants-of-motion}, it indicates that the number of constants of motion is 1 instead of the true number of 2 since a large jump exists from $n_c=1$ to $n_c=2$ and the training curve also presents a similar illusion.

According to the discussion above, we can see that although Meta-COMET and COMET have large differences in neural network architecture and training scheme, our method still retains the advantages of COMET. This inherited property makes our method equally or even more competitive in discovering the constant of motion since it is more lightweight and noise-robust than COMET.

\subsubsection{More complex cases}
In this subsection, two more complex cases are illustrated to demonstrate the general applicability of Meta-COMET.
\begin{figure}[t]
    \centering
    \begin{subfigure}{\linewidth}
        \centering
 \parbox{\textwidth} { \hspace{0.1cm} {\tiny COMET}  \hspace{1.4cm}  {\tiny Meta-COMET} \hspace{1.4cm} {\tiny True}} \\
        \includegraphics[width=\linewidth]{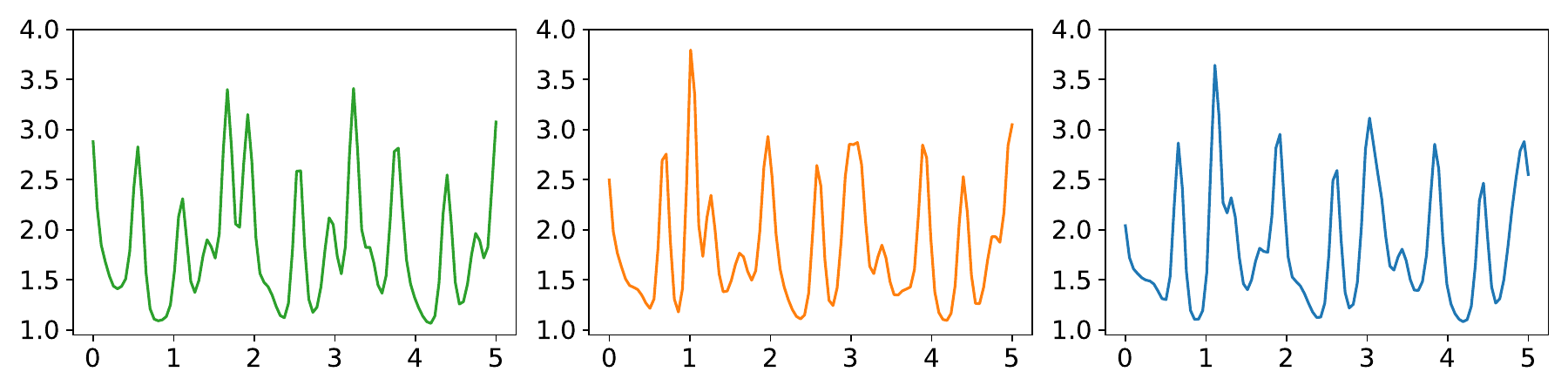}
    \end{subfigure}
    \begin{subfigure}{\linewidth}
        \flushleft
        \includegraphics[width=0.68\linewidth]{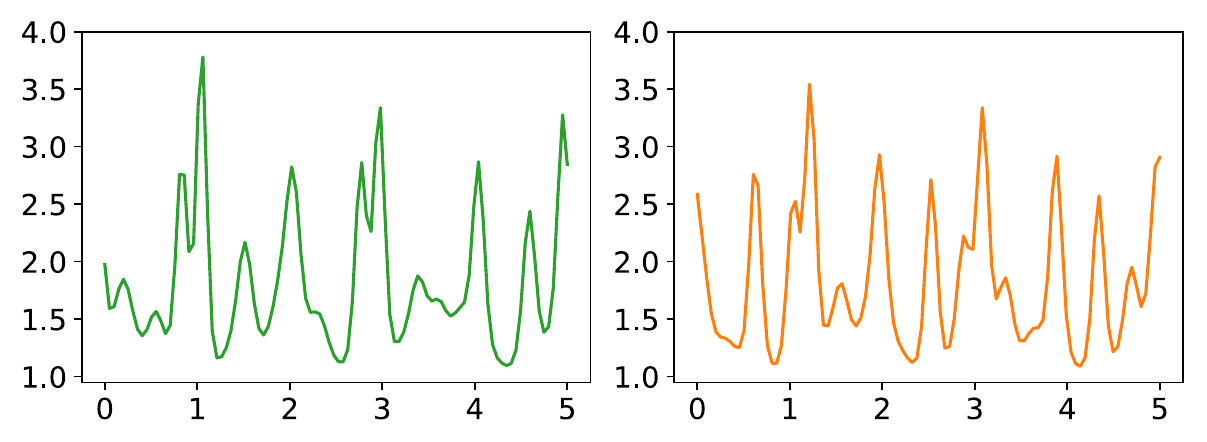}
    \end{subfigure}
        \begin{subfigure}{\linewidth}
        \flushleft
        \includegraphics[width=0.68\linewidth]{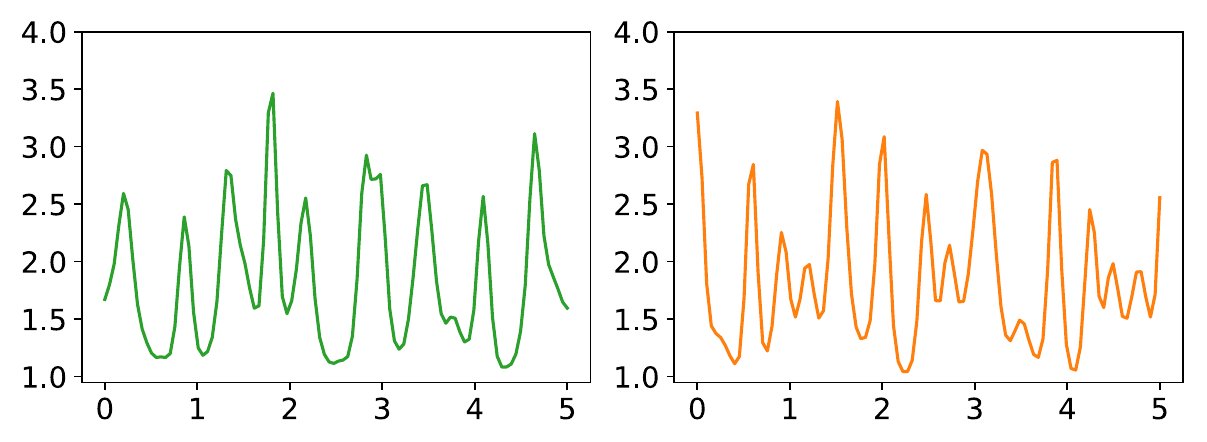}
    \end{subfigure}
    \caption{Plot of $u(x, t)$ at $t=20$ from simulations using COMET, Meta-COMET and true analytic expression with $\sigma = 0.05, 0.1, 0.2$ (from top to bottom).}
    \label{fig:kdv-trajectory}
\end{figure}
\label{sec:complex-cases}

\textbf{Korteweg-De Vries (KdV) equation} --- 
Different from previous cases, the KdV equation has an infinite number of states. The states, denoted by $u(x, t)$, are governed by the following equation: 
\begin{equation}
\frac{\partial u}{\partial t} = -u\frac{\partial u}{\partial x} - \delta^2 \frac{\partial^3 u}{\partial x^3}
\end{equation}
For numerical stability, we set $\delta = 0.00022$ in our case. The initial condition in the training dataset is periodic, expressed as $u(x, 0) = a_0 + a_1 \cos{(2\pi x/L + a_2)}$, where $a_0$, $a_1$, and $a_2$ are randomly sampled within the range of $[1.5, 2.5]$, $[0, 1]$, and $[0, 2\pi]$, respectively. The training dataset was generated by running 100 simulations from $t=0$ to $t=10$ with 100 steps, meanwhile sampling 100 evenly spaced points from $x=0$ to $x=L=5$.

Instead of two 1D convolutional hidden layers used in COMET, we also introduce \textsc{svd-FC} and \textsc{sd-FC} structures into the hidden layers of the neural networks utilized in this case. With that being thought, the $S$ and $D$ matrix are currently $hidden\textsc{-}dim \times r \times kernel\textsc{-}size$ matrixes, and $V$ is a $r \times kernel\textsc{-}size$ matrix. Each $kernel\textsc{-}size$ of matrix $S$ and $D$ are trained to be a semi-orthogonal matrix. We set $hidden\textsc{-}dim =250$, $r = 10$, and $kernel\textsc{-}size = 5$ in this case. In addition, We utilize circular padding in each layer, followed by the SiLU activation function except for the last one. The number of channels in the input is 1. And the output channel for 
$\mathbf{\dot{s}_0}$ and the density of $\mathbf{c}$, i.e. $\mathbf{c}_i = \int_{0}^{L} p_i(x)\ dx$ where $p_i(x)$ is the constant of motion density, are 1 and $n_c$, respectively. 

We compared the performance of Meta-COMET and COMET in solving the KdV equation for $t=0$ to $20$. Since the COMET method can take advantage of the constants of motion to refine the predictions, we set $n_c = 2$ and add different noise in the training dataset in this case. 
As shown in Figure~\ref{fig:kdv-trajectory}, all the experimental results show intact predictions. Moreover, when using Meta-COMET to solve the KdV equation, there are almost no cases where Scipy's \texttt{solve\_ivp} can not complete the integration within a reasonable time. Due to the complex evolution curve, it is not obvious which method achieves better results. Therefore, we calculate the 
average absolute deviation value, 
$\frac{\sum_{i=1}^{100}|u^i_p - u^i_{true}|}{100}$, between prediction $u^i_p$ and ground truth $u^i_{true}$, shown in table~\ref{tab:kdv_result}. It illustrates that although the number of parameters used in Meta-COMET is 56350, which is much less than COMET's 628251, the prediction results of our method deviate less from the true dynamics.

\begin{table}
\centering
\begin{center}
\begin{tabular}{lcc}
\hline
Cases & COMET & Meta-COMET   \\ 
\hline
$\sigma = 0.05$  & $ 0.423 $ & $ \mathbf{0.284}$   \\ [0.5ex]
\hline
$\sigma = 0.1$ & $ 0.628 $ & $ \mathbf{0.365}$  \\ [0.5ex]
\hline
$\sigma = 0.2$  & $ 0.725 $ & $ \mathbf{0.531}$   \\ [0.5ex]
\hline
\end{tabular}
\end{center}
\caption{Average deviation between predicted and true values}
\label{tab:kdv_result}
\end{table}

\textbf{Learning from pixels} --- 
\begin{figure*}[t]
    \centering
        \begin{subfigure}{\linewidth}
        \centering
        \includegraphics[width=\linewidth]{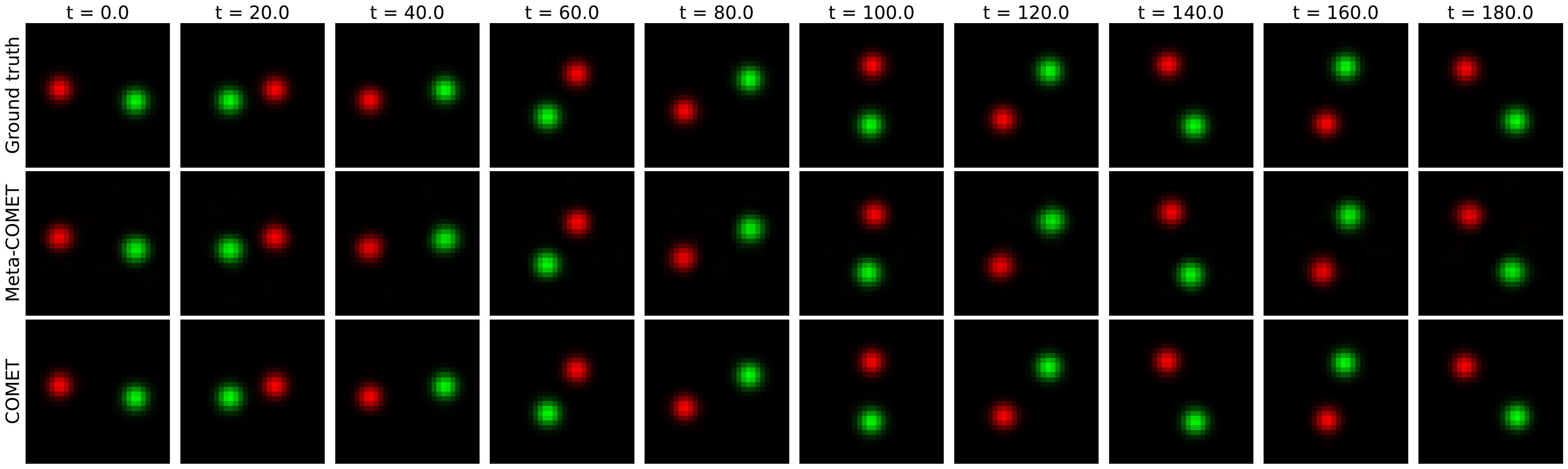}
        \caption{}
    \end{subfigure}
    \begin{subfigure}{\linewidth}
        \flushleft
        \includegraphics[width=\linewidth]{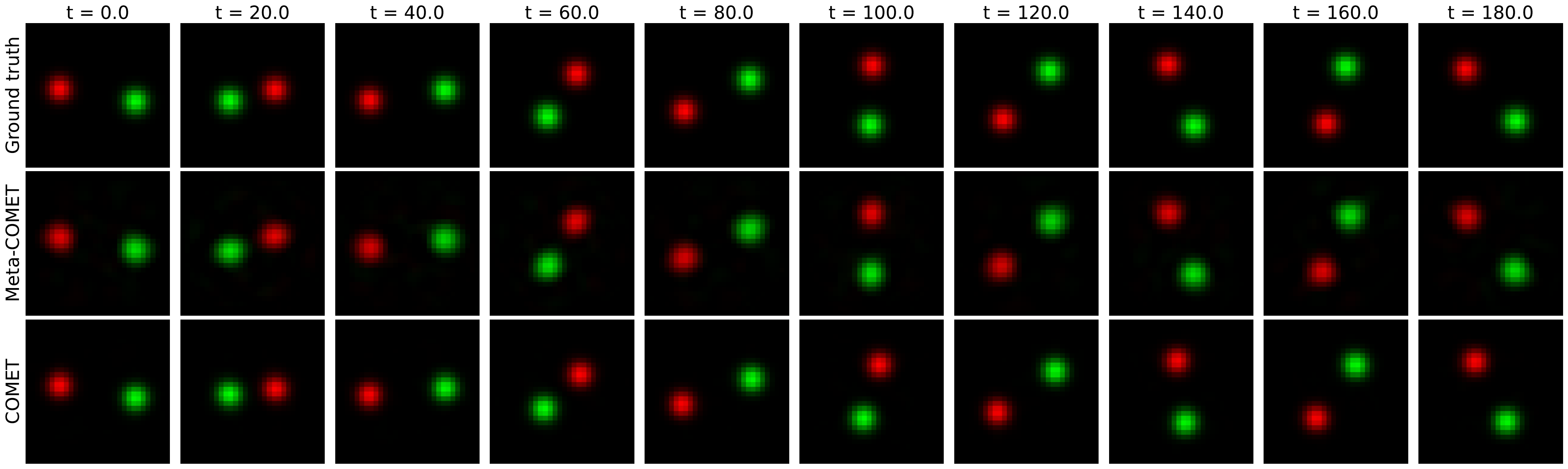}
        \caption{}
    \end{subfigure}
    \caption{Snapshots of the decoded images of the dynamics predicted by Meta-COMET, COMET as well as the ground truth images: (a) $\sigma = 0.0$, (b) $\sigma=0.001$ }
    \label{fig:vae-two-body}
\end{figure*}
In the second complex case, we conducted experiments on pixel data to show its capability to recover dynamics in latent space. Specifically, we simulated the dynamics of two bodies with gravitational interactions and presented the data as $30\times 30$ pixel images with 2 channels showing the positions of the two bodies.

Our model consists of an encoder, a Mate-COMET, and a decoder.
Two consecutive pixel frames were fed into the Mate-COMET model to represent the features about the position and velocity characteristics of the objects, expressed as $\mathbf{x}$. The encoder aims to convert $x$ to latent variables, $\mathbf{s}=\mathbf{f_e}(\mathbf{x})$. Following this, the Meta-COMET learns the dynamics of the latent variables, which will finally be converted back to pixel images, $\mathbf{\tilde{x}} = \mathbf{f_d}(\mathbf{s})$, by the decoder.

To reduce the number of parameters, aside from the Mate-COMET part, we also
introduce the \textsc{svd-FC} into the hidden layers of the encoder and decoder as well, denoted as $S_e, V_e, D_e$ and $S_d, V_d, D_d$. 
Due to this setting, 
except for the auto-encoder loss, i.e. $\mathcal{L}_{ae}=\lVert \mathbf{x} - \mathbf{\tilde{x}}\rVert^2$ and Meta-COMET loss, the loss function in Phase 1 should include a sum of semi-orthogonal constraint about $S_e, D_e$ and $S_d, D_d$. 
To make an approximation, we calculate the observed state derivatives $\hat{\dot{\mathbf{s}}}$ in equation~(\ref{eq:L2}) by finite difference of the encoded pixel data from two consecutive time steps, i.e. $\hat{\dot{\mathbf{s}}}\approx [\mathbf{f_e}(\mathbf{x}(t+\Delta t)) - \mathbf{f_e}(\mathbf{x}(t))] / \Delta t$, where $\Delta t = 0.01$.

In this experiment, we compared the performance of Meta-COMET and COMET, to learn the dynamics of the latent variables of the auto-encoder. The benchmark setup for the COMET model follows its original paper, but with 4 layers. For the auto-encoder and Meta-COMET in our model, we choose $r=100$, and SiLU activation between each layer in both parts. The number of latent states and constants of motion are set as 10 and 9. The total parameters in our model are 2163989, which is a reduction of approximately 4 percent compared to the COMET case. The orthogonal term in $\mathcal{L}_1$ is added into $\mathcal{L}_{2}$ to keep it smaller during optimization in this complex case.

Figure \ref{fig:vae-two-body} shows a sample of the decoded images of the dynamics predicted by COMET, and Meta-COMET. Compared to the ground truth, both methods can match the dynamics of the ground truth simulation until the end of the simulation from $t=0.0$ to $t=180$ in the absence of noise (Figure \ref{fig:vae-two-body}(a)). However, once the extra noise, $\sigma = 0.001$, is added to the latent state, COMET fails to mimic the true dynamics especially when $t=100$ or $t=140$. On the contrary, Meta-COMET can still complete the recovery task of the latent space (Figure \ref{fig:vae-two-body}(b)). This also reflects that our method is more robust to noise than COMET.

\section{Conclusions}

In conclusion, we propose the Meta-COMET approach in this paper. An elegant neural network architecture and a concise two-phase training algorithm are introduced. Extensive experiments demonstrate the superiority of our proposed method thanks to the delicately designed neural network architecture and training algorithm. Compared with COMET, our proposed method retains its advantages, such as freedom in coordinate selection and the possibility of finding the number of constants of motion, while being more lightweight and robust to noise. With these satisfactory results, we believe that Meta-COMET has great potential for scientific machine learning in the future.

However, with that being said, our results mainly focus on experimental comparisons and lack sufficient theoretical evidence of noise robustness. Hopefully, our work could stimulate more algorithmic and theoretical research to explore the properties of Meta-COMET and the constants of motion in dynamical systems.

\begin{acknowledgments}
This work was funded by NSFC, PR China under Grant No.12401676 and No.62372442.
\end{acknowledgments}


\bibliography{fwq_v16c}

\begin{thebibliography}{35}%
\makeatletter
\providecommand \@ifxundefined [1]{%
 \@ifx{#1\undefined}
}%
\providecommand \@ifnum [1]{%
 \ifnum #1\expandafter \@firstoftwo
 \else \expandafter \@secondoftwo
 \fi
}%
\providecommand \@ifx [1]{%
 \ifx #1\expandafter \@firstoftwo
 \else \expandafter \@secondoftwo
 \fi
}%
\providecommand \natexlab [1]{#1}%
\providecommand \enquote  [1]{``#1''}%
\providecommand \bibnamefont  [1]{#1}%
\providecommand \bibfnamefont [1]{#1}%
\providecommand \citenamefont [1]{#1}%
\providecommand \href@noop [0]{\@secondoftwo}%
\providecommand \href [0]{\begingroup \@sanitize@url \@href}%
\providecommand \@href[1]{\@@startlink{#1}\@@href}%
\providecommand \@@href[1]{\endgroup#1\@@endlink}%
\providecommand \@sanitize@url [0]{\catcode `\\12\catcode `\$12\catcode
  `\&12\catcode `\#12\catcode `\^12\catcode `\_12\catcode `\%12\relax}%
\providecommand \@@startlink[1]{}%
\providecommand \@@endlink[0]{}%
\providecommand \url  [0]{\begingroup\@sanitize@url \@url }%
\providecommand \@url [1]{\endgroup\@href {#1}{\urlprefix }}%
\providecommand \urlprefix  [0]{URL }%
\providecommand \Eprint [0]{\href }%
\providecommand \doibase [0]{http://dx.doi.org/}%
\providecommand \selectlanguage [0]{\@gobble}%
\providecommand \bibinfo  [0]{\@secondoftwo}%
\providecommand \bibfield  [0]{\@secondoftwo}%
\providecommand \translation [1]{[#1]}%
\providecommand \BibitemOpen [0]{}%
\providecommand \bibitemStop [0]{}%
\providecommand \bibitemNoStop [0]{.\EOS\space}%
\providecommand \EOS [0]{\spacefactor3000\relax}%
\providecommand \BibitemShut  [1]{\csname bibitem#1\endcsname}%
\let\auto@bib@innerbib\@empty
\bibitem [{\citenamefont {Noether}(1971)}]{noether1971invariant}%
  \BibitemOpen
  \bibfield  {author} {\bibinfo {author} {\bibfnamefont {E.}~\bibnamefont
  {Noether}},\ }\href@noop {} {\bibfield  {journal} {\bibinfo  {journal}
  {Transport theory and statistical physics}\ }\textbf {\bibinfo {volume}
  {1}},\ \bibinfo {pages} {186} (\bibinfo {year} {1971})}\BibitemShut {NoStop}%
\bibitem [{\citenamefont {Anderson}(1972)}]{anderson1972more}%
  \BibitemOpen
  \bibfield  {author} {\bibinfo {author} {\bibfnamefont {P.~W.}\ \bibnamefont
  {Anderson}},\ }\href@noop {} {\bibfield  {journal} {\bibinfo  {journal}
  {Science}\ }\textbf {\bibinfo {volume} {177}},\ \bibinfo {pages} {393}
  (\bibinfo {year} {1972})}\BibitemShut {NoStop}%
\bibitem [{\citenamefont {Elesedy}(2023)}]{elesedy2023symmetry}%
  \BibitemOpen
  \bibfield  {author} {\bibinfo {author} {\bibfnamefont {B.}~\bibnamefont
  {Elesedy}},\ }\emph {\bibinfo {title} {Symmetry and Generalisation in Machine
  Learning}},\ \href@noop {} {Ph.D. thesis},\ \bibinfo  {school} {PhD thesis,
  University of Oxford} (\bibinfo {year} {2023})\BibitemShut {NoStop}%
\bibitem [{\citenamefont {Sillerud}(2024)}]{sillerud2024space}%
  \BibitemOpen
  \bibfield  {author} {\bibinfo {author} {\bibfnamefont {L.~O.}\ \bibnamefont
  {Sillerud}},\ }in\ \href@noop {} {\emph {\bibinfo {booktitle} {Abiogenesis:
  The Physical Basis for Living Systems}}}\ (\bibinfo  {publisher} {Springer},\
  \bibinfo {year} {2024})\ pp.\ \bibinfo {pages} {129--209}\BibitemShut
  {NoStop}%
\bibitem [{\citenamefont {Sparavigna}(2015)}]{sparavigna2015historical}%
  \BibitemOpen
  \bibfield  {author} {\bibinfo {author} {\bibfnamefont {A.~C.}\ \bibnamefont
  {Sparavigna}},\ }\href@noop {} {\bibfield  {journal} {\bibinfo  {journal}
  {arXiv preprint arXiv:1511.07748}\ } (\bibinfo {year} {2015})}\BibitemShut
  {NoStop}%
\bibitem [{\citenamefont {Mukhamediev}\ \emph {et~al.}(2021)\citenamefont
  {Mukhamediev}, \citenamefont {Symagulov}, \citenamefont {Kuchin},
  \citenamefont {Yakunin},\ and\ \citenamefont
  {Yelis}}]{mukhamediev2021classical}%
  \BibitemOpen
  \bibfield  {author} {\bibinfo {author} {\bibfnamefont {R.~I.}\ \bibnamefont
  {Mukhamediev}}, \bibinfo {author} {\bibfnamefont {A.}~\bibnamefont
  {Symagulov}}, \bibinfo {author} {\bibfnamefont {Y.}~\bibnamefont {Kuchin}},
  \bibinfo {author} {\bibfnamefont {K.}~\bibnamefont {Yakunin}}, \ and\
  \bibinfo {author} {\bibfnamefont {M.}~\bibnamefont {Yelis}},\ }\href@noop {}
  {\bibfield  {journal} {\bibinfo  {journal} {Applied Sciences}\ }\textbf
  {\bibinfo {volume} {11}},\ \bibinfo {pages} {5541} (\bibinfo {year}
  {2021})}\BibitemShut {NoStop}%
\bibitem [{\citenamefont {Subramanian}\ \emph {et~al.}(2024)\citenamefont
  {Subramanian}, \citenamefont {Harrington}, \citenamefont {Keutzer},
  \citenamefont {Bhimji}, \citenamefont {Morozov}, \citenamefont {Mahoney},\
  and\ \citenamefont {Gholami}}]{subramanian2024towards}%
  \BibitemOpen
  \bibfield  {author} {\bibinfo {author} {\bibfnamefont {S.}~\bibnamefont
  {Subramanian}}, \bibinfo {author} {\bibfnamefont {P.}~\bibnamefont
  {Harrington}}, \bibinfo {author} {\bibfnamefont {K.}~\bibnamefont {Keutzer}},
  \bibinfo {author} {\bibfnamefont {W.}~\bibnamefont {Bhimji}}, \bibinfo
  {author} {\bibfnamefont {D.}~\bibnamefont {Morozov}}, \bibinfo {author}
  {\bibfnamefont {M.~W.}\ \bibnamefont {Mahoney}}, \ and\ \bibinfo {author}
  {\bibfnamefont {A.}~\bibnamefont {Gholami}},\ }\href@noop {} {\bibfield
  {journal} {\bibinfo  {journal} {Advances in Neural Information Processing
  Systems}\ }\textbf {\bibinfo {volume} {36}} (\bibinfo {year}
  {2024})}\BibitemShut {NoStop}%
\bibitem [{\citenamefont {Greydanus}\ \emph {et~al.}(2019)\citenamefont
  {Greydanus}, \citenamefont {Dzamba},\ and\ \citenamefont
  {Yosinski}}]{greydanus2019hamiltonian}%
  \BibitemOpen
  \bibfield  {author} {\bibinfo {author} {\bibfnamefont {S.}~\bibnamefont
  {Greydanus}}, \bibinfo {author} {\bibfnamefont {M.}~\bibnamefont {Dzamba}}, \
  and\ \bibinfo {author} {\bibfnamefont {J.}~\bibnamefont {Yosinski}},\
  }\href@noop {} {\bibfield  {journal} {\bibinfo  {journal} {Advances in Neural
  Information Processing Systems}\ }\textbf {\bibinfo {volume} {32}} (\bibinfo
  {year} {2019})}\BibitemShut {NoStop}%
\bibitem [{\citenamefont {Chen}\ \emph
  {et~al.}(2021{\natexlab{a}})\citenamefont {Chen}, \citenamefont {Matsubara},\
  and\ \citenamefont {Yaguchi}}]{chen2021nsf}%
  \BibitemOpen
  \bibfield  {author} {\bibinfo {author} {\bibfnamefont {Y.}~\bibnamefont
  {Chen}}, \bibinfo {author} {\bibfnamefont {T.}~\bibnamefont {Matsubara}}, \
  and\ \bibinfo {author} {\bibfnamefont {T.}~\bibnamefont {Yaguchi}},\
  }\href@noop {} {\bibfield  {journal} {\bibinfo  {journal} {Advances in Neural
  Information Processing Systems}\ }\textbf {\bibinfo {volume} {34}} (\bibinfo
  {year} {2021}{\natexlab{a}})}\BibitemShut {NoStop}%
\bibitem [{\citenamefont {Cranmer}\ \emph {et~al.}(2020)\citenamefont
  {Cranmer}, \citenamefont {Greydanus}, \citenamefont {Hoyer}, \citenamefont
  {Battaglia}, \citenamefont {Spergel},\ and\ \citenamefont
  {Ho}}]{cranmer2020lnn}%
  \BibitemOpen
  \bibfield  {author} {\bibinfo {author} {\bibfnamefont {M.}~\bibnamefont
  {Cranmer}}, \bibinfo {author} {\bibfnamefont {S.}~\bibnamefont {Greydanus}},
  \bibinfo {author} {\bibfnamefont {S.}~\bibnamefont {Hoyer}}, \bibinfo
  {author} {\bibfnamefont {P.}~\bibnamefont {Battaglia}}, \bibinfo {author}
  {\bibfnamefont {D.}~\bibnamefont {Spergel}}, \ and\ \bibinfo {author}
  {\bibfnamefont {S.}~\bibnamefont {Ho}},\ }\href@noop {} {\bibfield  {journal}
  {\bibinfo  {journal} {arXiv preprint arXiv:2003.04630}\ } (\bibinfo {year}
  {2020})}\BibitemShut {NoStop}%
\bibitem [{\citenamefont {Jin}\ \emph {et~al.}(2022)\citenamefont {Jin},
  \citenamefont {Zhang}, \citenamefont {Kevrekidis},\ and\ \citenamefont
  {Karniadakis}}]{jin2022learning-poisson}%
  \BibitemOpen
  \bibfield  {author} {\bibinfo {author} {\bibfnamefont {P.}~\bibnamefont
  {Jin}}, \bibinfo {author} {\bibfnamefont {Z.}~\bibnamefont {Zhang}}, \bibinfo
  {author} {\bibfnamefont {I.~G.}\ \bibnamefont {Kevrekidis}}, \ and\ \bibinfo
  {author} {\bibfnamefont {G.~E.}\ \bibnamefont {Karniadakis}},\ }\href@noop {}
  {\bibfield  {journal} {\bibinfo  {journal} {IEEE Transactions on Neural
  Networks and Learning Systems}\ } (\bibinfo {year} {2022})}\BibitemShut
  {NoStop}%
\bibitem [{\citenamefont {Zhang}\ \emph {et~al.}(2024)\citenamefont {Zhang},
  \citenamefont {Zhu},\ and\ \citenamefont {Lin}}]{zhang2024learning}%
  \BibitemOpen
  \bibfield  {author} {\bibinfo {author} {\bibfnamefont {J.}~\bibnamefont
  {Zhang}}, \bibinfo {author} {\bibfnamefont {Q.}~\bibnamefont {Zhu}}, \ and\
  \bibinfo {author} {\bibfnamefont {W.}~\bibnamefont {Lin}},\ }\href@noop {}
  {\bibfield  {journal} {\bibinfo  {journal} {Physical Review Research}\
  }\textbf {\bibinfo {volume} {6}},\ \bibinfo {pages} {L012031} (\bibinfo
  {year} {2024})}\BibitemShut {NoStop}%
\bibitem [{\citenamefont {Kasim}\ and\ \citenamefont
  {Lim}(2022)}]{kasim2022constants}%
  \BibitemOpen
  \bibfield  {author} {\bibinfo {author} {\bibfnamefont {M.~F.}\ \bibnamefont
  {Kasim}}\ and\ \bibinfo {author} {\bibfnamefont {Y.~H.}\ \bibnamefont
  {Lim}},\ }\href@noop {} {\bibfield  {journal} {\bibinfo  {journal} {Advances
  in Neural Information Processing Systems}\ }\textbf {\bibinfo {volume}
  {35}},\ \bibinfo {pages} {25295} (\bibinfo {year} {2022})}\BibitemShut
  {NoStop}%
\bibitem [{\citenamefont {Matsubara}\ and\ \citenamefont
  {Yaguchi}(2022)}]{matsubara2022finde}%
  \BibitemOpen
  \bibfield  {author} {\bibinfo {author} {\bibfnamefont {T.}~\bibnamefont
  {Matsubara}}\ and\ \bibinfo {author} {\bibfnamefont {T.}~\bibnamefont
  {Yaguchi}},\ }\href@noop {} {\bibfield  {journal} {\bibinfo  {journal} {arXiv
  preprint arXiv:2210.00272}\ } (\bibinfo {year} {2022})}\BibitemShut {NoStop}%
\bibitem [{\citenamefont {Ha}\ and\ \citenamefont
  {Jeong}(2021)}]{ha2021discovering}%
  \BibitemOpen
  \bibfield  {author} {\bibinfo {author} {\bibfnamefont {S.}~\bibnamefont
  {Ha}}\ and\ \bibinfo {author} {\bibfnamefont {H.}~\bibnamefont {Jeong}},\
  }\href@noop {} {\bibfield  {journal} {\bibinfo  {journal} {Physical Review
  Research}\ }\textbf {\bibinfo {volume} {3}},\ \bibinfo {pages} {L042035}
  (\bibinfo {year} {2021})}\BibitemShut {NoStop}%
\bibitem [{\citenamefont {Zhang}\ \emph {et~al.}(2023)\citenamefont {Zhang},
  \citenamefont {Weng}, \citenamefont {Das}, \citenamefont {Megretski},
  \citenamefont {Daniel},\ and\ \citenamefont {Nguyen}}]{zhang2023concernet}%
  \BibitemOpen
  \bibfield  {author} {\bibinfo {author} {\bibfnamefont {W.}~\bibnamefont
  {Zhang}}, \bibinfo {author} {\bibfnamefont {T.-W.}\ \bibnamefont {Weng}},
  \bibinfo {author} {\bibfnamefont {S.}~\bibnamefont {Das}}, \bibinfo {author}
  {\bibfnamefont {A.}~\bibnamefont {Megretski}}, \bibinfo {author}
  {\bibfnamefont {L.}~\bibnamefont {Daniel}}, \ and\ \bibinfo {author}
  {\bibfnamefont {L.~M.}\ \bibnamefont {Nguyen}},\ }in\ \href@noop {} {\emph
  {\bibinfo {booktitle} {International Conference on Machine Learning}}}\
  (\bibinfo {organization} {PMLR},\ \bibinfo {year} {2023})\ pp.\ \bibinfo
  {pages} {41694--41714}\BibitemShut {NoStop}%
\bibitem [{\citenamefont {Cooper}(2018)}]{cooper2018loss}%
  \BibitemOpen
  \bibfield  {author} {\bibinfo {author} {\bibfnamefont {Y.}~\bibnamefont
  {Cooper}},\ }\href@noop {} {\bibfield  {journal} {\bibinfo  {journal} {arXiv
  preprint arXiv:1804.10200}\ } (\bibinfo {year} {2018})}\BibitemShut {NoStop}%
\bibitem [{\citenamefont {Adeoye}\ \emph {et~al.}(2024)\citenamefont {Adeoye},
  \citenamefont {Petersen},\ and\ \citenamefont
  {Bemporad}}]{adeoye2024regularized}%
  \BibitemOpen
  \bibfield  {author} {\bibinfo {author} {\bibfnamefont {A.~D.}\ \bibnamefont
  {Adeoye}}, \bibinfo {author} {\bibfnamefont {P.~C.}\ \bibnamefont
  {Petersen}}, \ and\ \bibinfo {author} {\bibfnamefont {A.}~\bibnamefont
  {Bemporad}},\ }\href@noop {} {\bibfield  {journal} {\bibinfo  {journal}
  {arXiv preprint arXiv:2404.14875}\ } (\bibinfo {year} {2024})}\BibitemShut
  {NoStop}%
\bibitem [{\citenamefont {Cho}\ \emph {et~al.}(2024)\citenamefont {Cho},
  \citenamefont {Lee}, \citenamefont {Rim},\ and\ \citenamefont
  {Park}}]{cho2024hypernetwork}%
  \BibitemOpen
  \bibfield  {author} {\bibinfo {author} {\bibfnamefont {W.}~\bibnamefont
  {Cho}}, \bibinfo {author} {\bibfnamefont {K.}~\bibnamefont {Lee}}, \bibinfo
  {author} {\bibfnamefont {D.}~\bibnamefont {Rim}}, \ and\ \bibinfo {author}
  {\bibfnamefont {N.}~\bibnamefont {Park}},\ }\href@noop {} {\bibfield
  {journal} {\bibinfo  {journal} {Advances in Neural Information Processing
  Systems}\ }\textbf {\bibinfo {volume} {36}} (\bibinfo {year}
  {2024})}\BibitemShut {NoStop}%
\bibitem [{\citenamefont {Barraza}\ and\ \citenamefont
  {Gross}(2024)}]{barraza2024reduced}%
  \BibitemOpen
  \bibfield  {author} {\bibinfo {author} {\bibfnamefont {B.}~\bibnamefont
  {Barraza}}\ and\ \bibinfo {author} {\bibfnamefont {A.}~\bibnamefont
  {Gross}},\ }\href@noop {} {\bibfield  {journal} {\bibinfo  {journal}
  {Aerospace}\ }\textbf {\bibinfo {volume} {11}},\ \bibinfo {pages} {506}
  (\bibinfo {year} {2024})}\BibitemShut {NoStop}%
\bibitem [{\citenamefont {Epps}\ and\ \citenamefont
  {Krivitzky}(2019)}]{epps2019singular}%
  \BibitemOpen
  \bibfield  {author} {\bibinfo {author} {\bibfnamefont {B.~P.}\ \bibnamefont
  {Epps}}\ and\ \bibinfo {author} {\bibfnamefont {E.~M.}\ \bibnamefont
  {Krivitzky}},\ }\href@noop {} {\bibfield  {journal} {\bibinfo  {journal}
  {Experiments in Fluids}\ }\textbf {\bibinfo {volume} {60}},\ \bibinfo {pages}
  {1} (\bibinfo {year} {2019})}\BibitemShut {NoStop}%
\bibitem [{\citenamefont {Forrester}(2007)}]{forrester2007system}%
  \BibitemOpen
  \bibfield  {author} {\bibinfo {author} {\bibfnamefont {J.~W.}\ \bibnamefont
  {Forrester}},\ }\href@noop {} {\bibfield  {journal} {\bibinfo  {journal}
  {System Dynamics Review: The Journal of the System Dynamics Society}\
  }\textbf {\bibinfo {volume} {23}},\ \bibinfo {pages} {359} (\bibinfo {year}
  {2007})}\BibitemShut {NoStop}%
\bibitem [{\citenamefont {Tiumentsev}\ and\ \citenamefont
  {Egorchev}(2019)}]{tiumentsev2019neural}%
  \BibitemOpen
  \bibfield  {author} {\bibinfo {author} {\bibfnamefont {Y.}~\bibnamefont
  {Tiumentsev}}\ and\ \bibinfo {author} {\bibfnamefont {M.}~\bibnamefont
  {Egorchev}},\ }\href@noop {} {\emph {\bibinfo {title} {Neural network
  modeling and identification of dynamical systems}}}\ (\bibinfo  {publisher}
  {Academic Press},\ \bibinfo {year} {2019})\BibitemShut {NoStop}%
\bibitem [{\citenamefont {Chen}\ \emph {et~al.}(2018)\citenamefont {Chen},
  \citenamefont {Rubanova}, \citenamefont {Bettencourt},\ and\ \citenamefont
  {Duvenaud}}]{chen2018neural-ode}%
  \BibitemOpen
  \bibfield  {author} {\bibinfo {author} {\bibfnamefont {R.~T.}\ \bibnamefont
  {Chen}}, \bibinfo {author} {\bibfnamefont {Y.}~\bibnamefont {Rubanova}},
  \bibinfo {author} {\bibfnamefont {J.}~\bibnamefont {Bettencourt}}, \ and\
  \bibinfo {author} {\bibfnamefont {D.~K.}\ \bibnamefont {Duvenaud}},\
  }\href@noop {} {\bibfield  {journal} {\bibinfo  {journal} {Advances in neural
  information processing systems}\ }\textbf {\bibinfo {volume} {31}} (\bibinfo
  {year} {2018})}\BibitemShut {NoStop}%
\bibitem [{\citenamefont {Greydanus}\ and\ \citenamefont
  {Sosanya}(2022)}]{greydanus2022dissipative-hnn}%
  \BibitemOpen
  \bibfield  {author} {\bibinfo {author} {\bibfnamefont {S.}~\bibnamefont
  {Greydanus}}\ and\ \bibinfo {author} {\bibfnamefont {A.}~\bibnamefont
  {Sosanya}},\ }\href@noop {} {\bibfield  {journal} {\bibinfo  {journal} {arXiv
  preprint arXiv:2201.10085}\ } (\bibinfo {year} {2022})}\BibitemShut {NoStop}%
\bibitem [{\citenamefont {Zhong}\ \emph {et~al.}(2020)\citenamefont {Zhong},
  \citenamefont {Dey},\ and\ \citenamefont
  {Chakraborty}}]{zhong2020dissipative}%
  \BibitemOpen
  \bibfield  {author} {\bibinfo {author} {\bibfnamefont {Y.~D.}\ \bibnamefont
  {Zhong}}, \bibinfo {author} {\bibfnamefont {B.}~\bibnamefont {Dey}}, \ and\
  \bibinfo {author} {\bibfnamefont {A.}~\bibnamefont {Chakraborty}},\
  }\href@noop {} {\bibfield  {journal} {\bibinfo  {journal} {arXiv preprint
  arXiv:2002.08860}\ } (\bibinfo {year} {2020})}\BibitemShut {NoStop}%
\bibitem [{\citenamefont {Han}\ \emph {et~al.}(2021)\citenamefont {Han},
  \citenamefont {Glaz}, \citenamefont {Haile},\ and\ \citenamefont
  {Lai}}]{han2021adaptable}%
  \BibitemOpen
  \bibfield  {author} {\bibinfo {author} {\bibfnamefont {C.-D.}\ \bibnamefont
  {Han}}, \bibinfo {author} {\bibfnamefont {B.}~\bibnamefont {Glaz}}, \bibinfo
  {author} {\bibfnamefont {M.}~\bibnamefont {Haile}}, \ and\ \bibinfo {author}
  {\bibfnamefont {Y.-C.}\ \bibnamefont {Lai}},\ }\href@noop {} {\bibfield
  {journal} {\bibinfo  {journal} {Physical Review Research}\ }\textbf {\bibinfo
  {volume} {3}},\ \bibinfo {pages} {023156} (\bibinfo {year}
  {2021})}\BibitemShut {NoStop}%
\bibitem [{\citenamefont {Finzi}\ \emph {et~al.}(2020)\citenamefont {Finzi},
  \citenamefont {Wang},\ and\ \citenamefont {Wilson}}]{finzi2020simplifying}%
  \BibitemOpen
  \bibfield  {author} {\bibinfo {author} {\bibfnamefont {M.}~\bibnamefont
  {Finzi}}, \bibinfo {author} {\bibfnamefont {K.~A.}\ \bibnamefont {Wang}}, \
  and\ \bibinfo {author} {\bibfnamefont {A.~G.}\ \bibnamefont {Wilson}},\
  }\href@noop {} {\bibfield  {journal} {\bibinfo  {journal} {Advances in neural
  information processing systems}\ }\textbf {\bibinfo {volume} {33}},\ \bibinfo
  {pages} {13880} (\bibinfo {year} {2020})}\BibitemShut {NoStop}%
\bibitem [{\citenamefont {Chen}\ \emph {et~al.}(2019)\citenamefont {Chen},
  \citenamefont {Zhang}, \citenamefont {Arjovsky},\ and\ \citenamefont
  {Bottou}}]{chen2019symplectic-rnn}%
  \BibitemOpen
  \bibfield  {author} {\bibinfo {author} {\bibfnamefont {Z.}~\bibnamefont
  {Chen}}, \bibinfo {author} {\bibfnamefont {J.}~\bibnamefont {Zhang}},
  \bibinfo {author} {\bibfnamefont {M.}~\bibnamefont {Arjovsky}}, \ and\
  \bibinfo {author} {\bibfnamefont {L.}~\bibnamefont {Bottou}},\ }in\
  \href@noop {} {\emph {\bibinfo {booktitle} {International Conference on
  Learning Representations}}}\ (\bibinfo {year} {2019})\BibitemShut {NoStop}%
\bibitem [{\citenamefont {Zhong}\ \emph {et~al.}(2019)\citenamefont {Zhong},
  \citenamefont {Dey},\ and\ \citenamefont {Chakraborty}}]{zhong2019sympnet}%
  \BibitemOpen
  \bibfield  {author} {\bibinfo {author} {\bibfnamefont {Y.~D.}\ \bibnamefont
  {Zhong}}, \bibinfo {author} {\bibfnamefont {B.}~\bibnamefont {Dey}}, \ and\
  \bibinfo {author} {\bibfnamefont {A.}~\bibnamefont {Chakraborty}},\ }in\
  \href@noop {} {\emph {\bibinfo {booktitle} {International Conference on
  Learning Representations}}}\ (\bibinfo {year} {2019})\BibitemShut {NoStop}%
\bibitem [{\citenamefont {Chen}\ \emph
  {et~al.}(2021{\natexlab{b}})\citenamefont {Chen}, \citenamefont {Matsubara},\
  and\ \citenamefont {Yaguchi}}]{chen2021neural}%
  \BibitemOpen
  \bibfield  {author} {\bibinfo {author} {\bibfnamefont {Y.}~\bibnamefont
  {Chen}}, \bibinfo {author} {\bibfnamefont {T.}~\bibnamefont {Matsubara}}, \
  and\ \bibinfo {author} {\bibfnamefont {T.}~\bibnamefont {Yaguchi}},\
  }\href@noop {} {\bibfield  {journal} {\bibinfo  {journal} {Advances in Neural
  Information Processing Systems}\ }\textbf {\bibinfo {volume} {34}},\ \bibinfo
  {pages} {16659} (\bibinfo {year} {2021}{\natexlab{b}})}\BibitemShut {NoStop}%
\bibitem [{\citenamefont {Lutter}\ \emph {et~al.}(2019)\citenamefont {Lutter},
  \citenamefont {Ritter},\ and\ \citenamefont {Peters}}]{lutter2019delan}%
  \BibitemOpen
  \bibfield  {author} {\bibinfo {author} {\bibfnamefont {M.}~\bibnamefont
  {Lutter}}, \bibinfo {author} {\bibfnamefont {C.}~\bibnamefont {Ritter}}, \
  and\ \bibinfo {author} {\bibfnamefont {J.}~\bibnamefont {Peters}},\
  }\href@noop {} {\bibfield  {journal} {\bibinfo  {journal} {arXiv preprint
  arXiv:1907.04490}\ } (\bibinfo {year} {2019})}\BibitemShut {NoStop}%
\bibitem [{\citenamefont {Ryu}\ \emph {et~al.}(2024)\citenamefont {Ryu},
  \citenamefont {Park}, \citenamefont {Lee},\ and\ \citenamefont
  {Choi}}]{ryu2024physics}%
  \BibitemOpen
  \bibfield  {author} {\bibinfo {author} {\bibfnamefont {I.}~\bibnamefont
  {Ryu}}, \bibinfo {author} {\bibfnamefont {G.-B.}\ \bibnamefont {Park}},
  \bibinfo {author} {\bibfnamefont {Y.}~\bibnamefont {Lee}}, \ and\ \bibinfo
  {author} {\bibfnamefont {D.-H.}\ \bibnamefont {Choi}},\ }\href@noop {}
  {\bibfield  {journal} {\bibinfo  {journal} {Journal of Mechanical Science and
  Technology}\ ,\ \bibinfo {pages} {1}} (\bibinfo {year} {2024})}\BibitemShut
  {NoStop}%
\bibitem [{\citenamefont {Householder}(1958)}]{householder1958unitary}%
  \BibitemOpen
  \bibfield  {author} {\bibinfo {author} {\bibfnamefont {A.~S.}\ \bibnamefont
  {Householder}},\ }\href@noop {} {\bibfield  {journal} {\bibinfo  {journal}
  {Journal of the ACM (JACM)}\ }\textbf {\bibinfo {volume} {5}},\ \bibinfo
  {pages} {339} (\bibinfo {year} {1958})}\BibitemShut {NoStop}%
\bibitem [{\citenamefont {Golub}\ and\ \citenamefont
  {Van~Loan}(2013)}]{golub2013matrix}%
  \BibitemOpen
  \bibfield  {author} {\bibinfo {author} {\bibfnamefont {G.~H.}\ \bibnamefont
  {Golub}}\ and\ \bibinfo {author} {\bibfnamefont {C.~F.}\ \bibnamefont
  {Van~Loan}},\ }\href@noop {} {\bibfield  {journal} {\bibinfo  {journal}
  {Johns Hopkins}\ } (\bibinfo {year} {2013})}\BibitemShut {NoStop}%
\end{thebibliography}%


\end{document}